\documentclass[10pt,twocolumn,letterpaper]{article}

\usepackage{iccv}
\usepackage{times}
\usepackage{epsfig}
\usepackage{graphicx}
\usepackage{subfigure}
\usepackage{amsmath}
\usepackage{amssymb}
\usepackage{multirow}
\usepackage{caption}
\graphicspath{{figs/}}
\newenvironment{packed_itemize}{
	\begin{itemize}
		\setlength{\itemsep}{1pt}
		\setlength{\parskip}{0pt}
		\setlength{\parsep}{0pt}
	}{\end{itemize}}
\usepackage{bm}

\usepackage[T1]{fontenc}
\usepackage[utf8]{inputenc}
\usepackage{authblk}

\usepackage[pagebackref=true,breaklinks=true,colorlinks,urlcolor=blue,bookmarks=false]{hyperref}

\iccvfinalcopy 

\def\iccvPaperID{3816} 

\ificcvfinal\fi

\begin{document}

\title{Geometry-Aware Self-Training for Unsupervised Domain Adaptation\\ on Object Point Clouds}
\author{Longkun Zou$^{1,2}$} 
\author{Hui Tang$^1$}
\author{Ke Chen$^{1,3*}$}
\author{Kui Jia$^{1,3,4}$\thanks{Corresponding authors}}
\affil{$^1$South China University of Technology, $^2$DexForce Technology Co., Ltd\\ $^3$Peng Cheng Laboratory, $^4$Pazhou Laboratory}
\affil{\tt\small \{eelongkunzou, eehuitang\}@mail.scut.edu.cn, \{chenk, kuijia\}@scut.edu.cn}


\maketitle
\ificcvfinal\thispagestyle{empty}\fi

\begin{abstract}
The point cloud representation of an object can have a large geometric variation in view of inconsistent data acquisition procedure, which thus leads to domain discrepancy due to diverse and uncontrollable shape representation cross datasets. To improve discrimination on unseen distribution of point-based geometries in a practical and feasible perspective, this paper proposes a new method of geometry-aware self-training (GAST) for unsupervised domain adaptation of object point cloud classification. Specifically, this paper aims to learn a domain-shared representation of semantic categories, via two novel self-supervised geometric learning tasks as feature regularization.
On one hand, the representation learning is empowered by a linear mixup of point cloud samples with their self-generated rotation labels, to capture a global topological configuration of local geometries. 
On the other hand, a diverse point distribution across datasets can be normalized with a novel curvature-aware distortion localization. 
Experiments on the PointDA-10 dataset show that our GAST method can significantly outperform the state-of-the-art methods. Source codes and pre-trained models are available at \url{https://github.com/zou-longkun/GAST}.\vspace{-0.5cm}
\end{abstract}

	\section{Introduction}
	\label{SecIntro}
	
	The point cloud is a popular shape representation widely adopted in 3D object classification \cite{Pointnet,DGCNN,Pointnet++,tang2020improving}, owing to its simple structure and easy acquisition.
	Specifically, point clouds can be generated via point sampling on the surface of object models, which is the recent typical solution to generate synthetic datasets for point cloud classification, \eg the ShapeNet \cite{Shapenet} and ModelNet \cite{Modelnet} benchmarks. 
	Beyond synthetic point clouds, a large-scale size of point clouds as a raw output of popular 3D sensors such as LiDAR and depth cameras can be collected in practice.
	Synthetic point clouds of each dataset typically follow a strategy, \eg uniformly sampling over the whole object surface in the ShapeNet \cite{Shapenet}, and thus are under the controllable generation procedure.
	Point clouds existing in real world have a large geometric variation in view of the existence of realistic sensor noises, non-uniform point distribution, and single-view coverage of unclosed surface due to self-occlusion. 
	In view of this, point-based shape representation can have the shifts of distribution, which desires domain adaptation techniques to improve generalization of point cloud classifiers.
	
	\begin{figure}[t]
		\centering
		\includegraphics[width=0.8\linewidth]{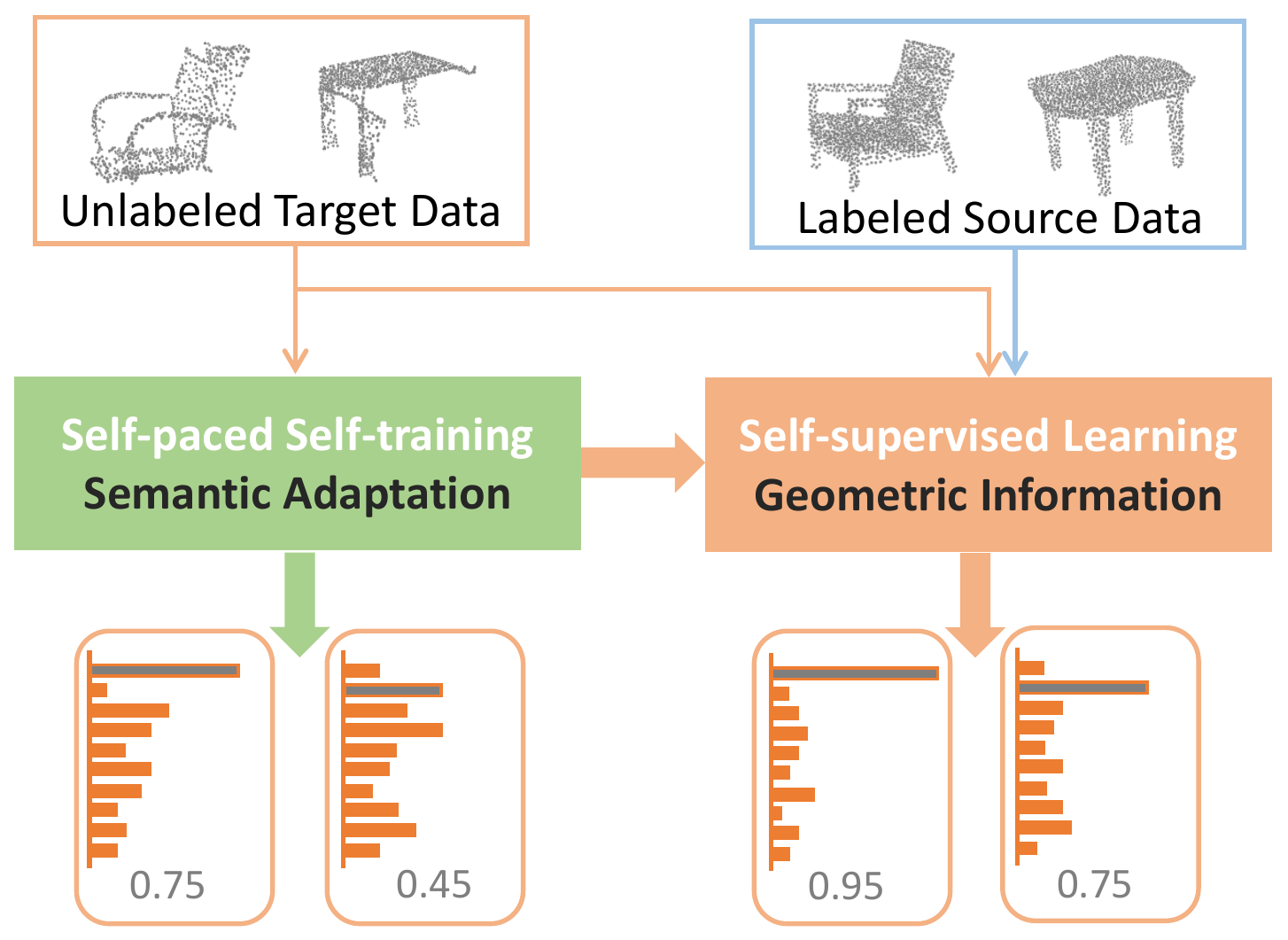}\vspace{-0.2cm}
		\caption{Concept of our geometry-aware self-training UDA on point cloud classification. {We compare conventional self-paced self-training semantic adaptation (highlighted in green) and the proposed geometry-aware self-training (with both blocks).
			Note that the gray numbers in the bottom are predicted classification probabilities for target testing point clouds from the ScanNet \cite{Scannet} (\ie the two target examples on the top left), which can verify the effectiveness of our self-supervised geometry-aware feature regularization on semnatic representation. }
		} \label{fig.intro}\vspace{-0.6cm}
	\end{figure}
	
	On one hand, a large number of synthetic point clouds can be readily generated based on object CAD models with corresponding semantic labels, which thus leads to sufficient labeled point cloud samples.
	On the other hand, real point clouds typically demand expensive manual annotations, which therefore causes a limited size of real data.
	In a practical perspective, a promising setting of domain adaptation on point cloud classification is to leverage the data from a label-rich source domain (\eg synthetic point clouds) with mining certain inter-relation to the target task on a label-scarce target domain (\eg real point clouds). 
	Motivated by the above observation, this paper concerns on unsupervised domain adaptation (UDA) \cite{transfer_learning_survey} of object point clouds, \ie to cope with the data distribution discrepancy  \cite{da_theory2,da_theory1} of point-based shape representation.
	Such a problem aims to learn a model with both training samples from labeled source and unlabeled target point sets that can classify target testing samples into one of the common semantic categories of two domains.
	
	UDA for 2D image classification \cite{dan,can,dann,mcd,mstn,tpn} has been well investigated for years based on domain adaptation theories \cite{da_theory2,da_theory1}, while very few works \cite{PointDAN,PCM_RegRecT} explore UDA for point cloud classification.
	Specifically, these UDA methods on point clouds concern on either semantic feature adaptation via explicit feature alignment across domains as existing image-based UDA \cite{PointDAN} or self-supervised feature encoding for domain-invariant geometric patterns without bridging the domain gap of semantic features \cite{PCM_RegRecT}, resulting in a sub-optimal adaptation. 
	Encouraged by the state-of-the-art self-training method \cite{cbst}, this paper adopts a self-paced self-training (SPST) scheme as our baseline to incorporate target discrimination into the semantic representation, via discovering structural similarity of inter-domain semantic patterns.
However, such a SPST method is directly adapted from 2D UDA domain, which omits inherent geometric ambiguities of point cloud representations. 
	
In view of recent success of self-supervised learning on point clouds \cite{SSL_UDA_Images,SSL_RotPred_PC,SSL_PointSeg_PC,tang2020improving} to incorporate local or global geometries to semantic feature representation, this paper proposes
a novel \emph{Geometry-Aware Self-Training (GAST)} method for UDA on point clouds, which designs two simple yet effective self-supervised tasks to regularize semantic feature encoding beyond the SPST baseline,  whose concept is also illustrated in Figure \ref{fig.intro}.
Specifically, this paper introduces 1) a point cloud mixup for rotation angle classification to discover objects' global topological structure; and 2) curvature-aware distortion localization for feature robustness against inconsistent point distribution.
As the source and target point clouds do not have supervision signals for the two pretext tasks, samples from both domains with self-generated rotation/location index labels can be trained jointly in a supervised style.
Consequently, geometric patterns captured by self-supervised tasks are shared between both domains, which thus can further boost discrimination of semantic representation to classify target point clouds. 
	Experiments on the 3D UDA benchmarking PointDA-10 \cite{PointDAN} show the superiority of our proposed method over the state-of-the-art methods significantly. Our contributions are summarized as follows.
	\begin{packed_itemize}
		\item This paper proposes a novel \emph{Geometry-Aware Self-Training} method for unsupervised domain adaptation on object point sets, which encodes domain-invariant geometrics to semantic representation to mitigate domain discrepancy of point-based representations. 
		\item Technically, based on self-paced self-training on unlabeled target data, our GAST integrates the self-supervised tasks of predicting rotation class and distortion location into representation learning, such that the domain-shared feature space can be constructed. 
		\item Experiments on the public benchmark verify that the proposed GAST achieves the new state-of-the-art performance of unsupervised domain adaption on point cloud classification, especially performs consistently the best for the more important synthetic-to-real tasks.
	\end{packed_itemize}

	\section{Related Work}
	\label{SecRelatedWork}
	
	\vspace{0.1cm}\noindent\textbf{Deep Classification on Point Clouds --} 
	Most of recent point cloud classification networks \cite{Pointnet,DGCNN,Pointnet++,tang2020improving} concern on coping with sparsity and irregularity of point-based shape representation, which can be categorized into two groups. 
	The first group of algorithms \cite{SO_Net, ModelingPC,Pointweb,StructuralRR,PointASNLRP} is designed based on multi-layer perceptron (MLP), which densely encode features on each point independently to aggregate a global shape representation.
	The second group of algorithms \cite{DGCNN,SphericalC,Clusternet} aims to encoding each point's local neighborhood into feature representation, either by constructing a spatial/spectral graph \cite{DGCNN,Clusternet} or by defining Euclidean convolution operation on irregular points with a continuous space (\eg a sphere \cite{SphericalC,LearningSE}) or regular grids (\eg voxels) \cite{KPconv,PointCNN}.  
	These methods for point cloud classification attempt to learn discriminative semantic features from global and/or local geometries, but very few work \cite{PointDAN,PCM_RegRecT} pay attention to mitigating distribution shifts of point-based representation, which is our main concern in this paper.

	\begin{figure*}[t]
		\centering
		\includegraphics[width=0.75\linewidth]{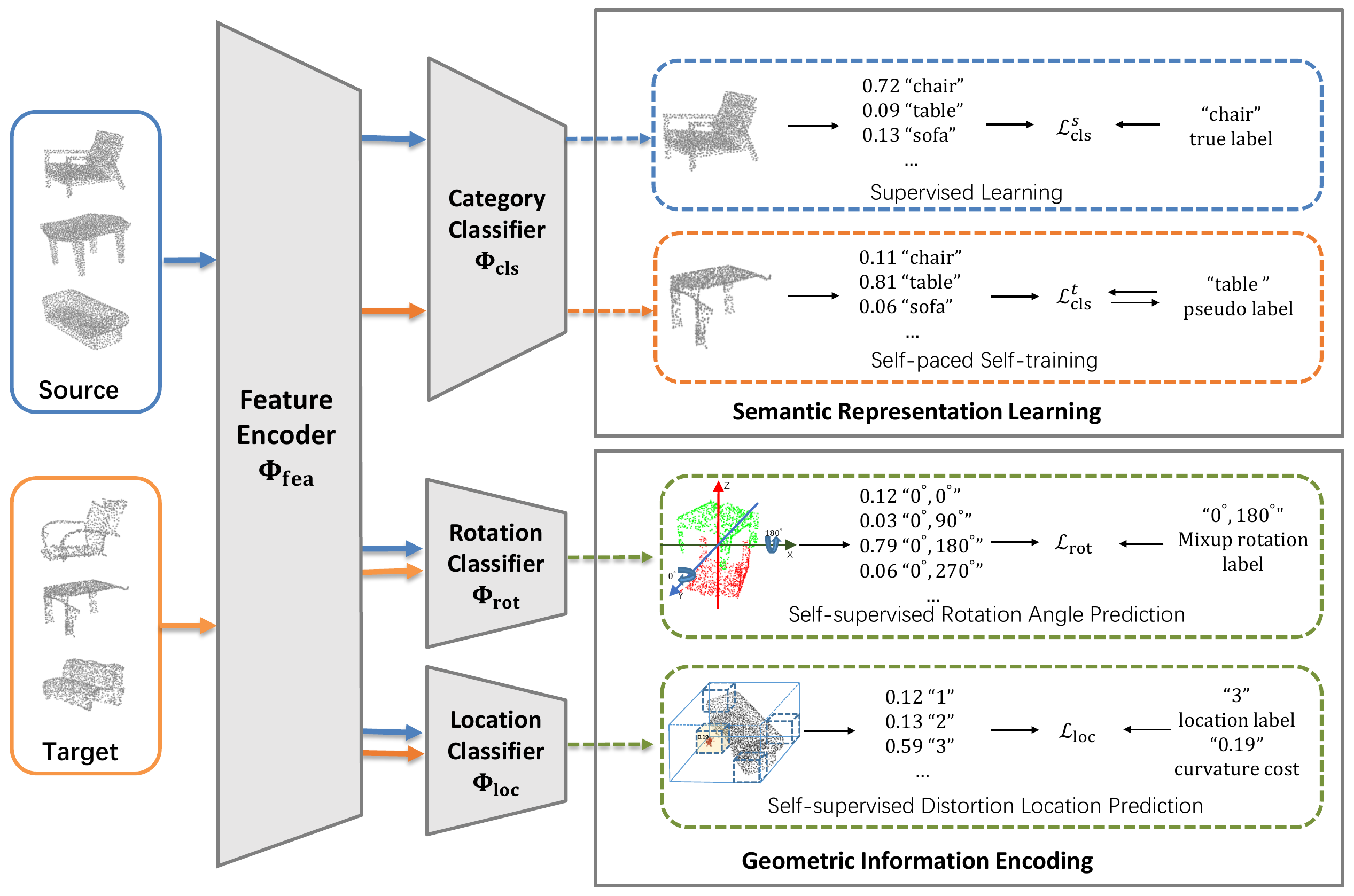}\vspace{-0.2cm}
		\caption{Overview of our \emph{Geometry-Aware Self-Training}, which includes four key components: supervised training on source domain, self-paced self-training on target domain, self-supervised rotation angle prediction, and self-supervised distortion location prediction, corresponding to losses of $\mathcal{L}_\text{cls}^s$, $\mathcal{L}_\text{cls}^t$, $\mathcal{L}_\text{rot}$, and $\mathcal{L}_\text{loc}$. The three individual classifiers take the features from the shared feature encoder as their input. Note that the small black arrow indicates the direction of optimization.}
		\label{Fig.Architecture}\vspace{-0.6cm}
	\end{figure*}
	
	\vspace{0.1cm}\noindent\textbf{2D and 3D Unsupervised Domain Adaptation--}
	Mainstream UDA methods for 2D image classification differ mainly in the strategy of reducing the discrepancy across domains and are accordingly divided into two categories. 
	Methods in the first category minimize a proxy of the domain discrepancy, which is measured by distribution statistics \cite{dan,can,BeyondSW} or distance metrics \cite{mstn,tpn,cat}. 
	The second category includes methods aligning source and target feature distributions in an adversarial manner, \ie playing minimax games \cite{gans} at the levels of domain \cite{dann,adda,SimNet} or category \cite{mcd,adr,swd,bsp}. 
	Recently, a few works \cite{PointDAN,PCM_RegRecT} propose the problem of UDA on irregular point-based representation, which inherits the challenge of semantic gap as other DA problems and also has its specific challenge of domain-agnostic feature encoding from local geometries of point clouds. Qin \etal \cite{PointDAN} propose a node module with adaptive receptive field to model the discriminative local structures and minimize an MMD loss to explicitly align local features across domains. 
	Achituve \etal \cite{PCM_RegRecT} learn an informative representation with abundant local geometrics in a self-supervised manner, \ie reconstructing a partially distorted point cloud. 
	Their limitations as discussed in Sec. \ref{SecIntro} encourage us to propose our GAST method for a discriminative domain-shared representation. Experiments in Table \ref{table:PointDA10_comparison} can verify superior performance of GAST to PointDAN \cite{PointDAN} and DefRec \cite{PCM_RegRecT}.

	\vspace{0.1cm}\noindent\textbf{Self-supervised Learning --}
	Self-supervised learning leverages the input itself as supervision for a pretext task, which learns the representation benefiting downstream tasks. 
	A comprehensive summary of existing methods in this direction is provided by \cite{SSL_Survey} and we retrospect those most related ones. 
	For learning geometric feature from point clouds, Sauder \etal \cite{SSL_PointSeg_PC} design to split an input point cloud into several parts with a random permutation on these parts and the goal is to predict the original permutation, while Poursaeed \etal \cite{SSL_RotPred_PC} propose to rotate the whole input and predict the rotation angle. 
	In this work, our GAST adopts two pretext tasks -- the task of rotation angle prediction similar to \cite{PCM_RegRecT} and the novel distortion localization to distinguish the distorted part from other parts, both of which are formulated into a classification problem.
	Effects of two pretext tasks are evaluated in Table \ref{table:PointDA10_comparison}, which can verify our motivation.
	
	
	\section{Methodology}\label{SecMethod}
	
	In unsupervised domain adaptation (UDA) on point sets, given a source domain $\mathcal{S} = \{\mathcal{P}_i^s, y_i^s\}_{i=1}^{n_s}$ with $n_s$ labeled samples and a target domain $\mathcal{T} = \{\mathcal{P}_i^t\}_{i=1}^{n_t}$ with $n_t$ unlabeled samples, a semantic label space $\mathcal{Y}$ is shared between $\mathcal{S}$ and $\mathcal{T}$ (\ie $\mathcal{Y}_{s} = \mathcal{Y}_{t}$), where point cloud $\mathcal{P} \in \mathcal{X} \subset \mathbb{R}^{m \times 3}$ consisting of $m$ three-dimensional coordinate points $(x,y,z)$ represents one object shape.
	Let the number of categories $|\mathcal{Y}|$ be $C$, \ie $y^s \in \{1, 2, \cdots, C\}$ for any source instance $\mathcal{P}^s$. 
	The objective of point-based UDA is to learn a domain-adapted mapping function $\Phi: \mathcal{X} \rightarrow \mathcal{Y}$ that can correctly classify point cloud samples into one of $C$ semantic categories.
	In the context of deep learning, the mapping function $\Phi$ can be formulated into a cascade of a feature encoder $\Phi_\text{fea}: \mathcal{X} \rightarrow \mathbb{R}^d$ for any input $\mathcal{P}$ and a classifier $\Phi_\text{cls}: \mathbb{R}^{d}\rightarrow [0,1]^C$ typically using fully-connected layers as follows:
	\begin{equation}
	\Phi(\mathcal{P}) = \Phi_\text{cls}(\bm{z}) \circ \Phi_\text{fea}(\mathcal{P})
	\end{equation}
	where $d$ denotes the dimension of the feature representation output $\bm{z} \in \mathcal{Z}$ of $\Phi_\text{fea}(\mathcal{P})$. 
	Denote the category probability vector of $\mathcal{P}$ as $\bm{p}=\Phi_\text{cls}(\bm{z})=[p_1, \cdots, p_C]$ subject to $\sum_{i=1}^C p_i = 1$.
	Since both $\mathcal{S}$ and $\mathcal{T}$ domains by assumption follow different data distributions, the main challenge in point-based UDA is to reduce domain discrepancy of feature encoding $\Phi_\text{fea}$ in terms of semantics and geometrics.
	
	This paper introduces a novel geometry-aware self-training (GAST) method for UDA on point set classification, whose pipeline is illustrated in Figure \ref{Fig.Architecture}.
	Specifically, the proposed GAST method is made up of two parts -- semantic feature adaptation (see Sec. \ref{SecSemAda}) and geometry-aware regularization (see Sec. \ref{SecGeoEnc}). 
	Without explicit feature alignment, our GAST applies iterative self-training with self-paced learning on target data to adapt semantic representation, which is extracted from source data in a supervised learning style. 
	To complement the semantic representation learning, the proposed GAST regularizes feature learning via incorporating global and local geometric structures by self-supervision of predicting rotation angle and distortion location.

	\subsection{Self-Paced Semantic Feature Adaptation}
	\label{SecSemAda}
	
	We aim to learn an adaptive classification model for generalizing knowledge induced from a labeled source domain to an unlabeled target one.
	As discussed before, due to domain discrepancy, the semantic representation generated by learning the classification model $\Phi_\text{cls} \circ \Phi_\text{fea}$ on source data with category labels $y$, can lead to classification performance degrading significantly when applied to the instances of unlabeled target domain. 
	In this way, we propose to learn the domain-shared semantic representation via training the same network $\Phi_\text{cls} \circ \Phi_\text{fea}$ with source and target samples jointly. 
	For training with unlabeled target samples in a supervised learning method, we adopt the self-training scheme in a self-paced learning manner to optimally select confident target samples, which together with pseudo labels are fed into the classification model $\Phi$ to refine semantic feature with target discrimination. 
	
	\vspace{0.1cm}\noindent\textbf{Supervised Learning on Source Domain --}
	Denote the labeled source samples $\{\mathcal{P}_i^s, y_i^s\}_{i=1}^{n_s}$ and their category probability vectors $\{\bm{p}_i^s\}_{i=1}^{n_s}$ predicted by the model $\Phi$, which is trained via minimizing the cross-entropy loss:
	\begin{eqnarray}\label{EqnSrcSupLearn}
	\begin{aligned}
	\min_{\Phi_\text{fea},\Phi_\text{cls}} \mathcal{L}_\text{cls}^s = - \frac{1}{n_s} \sum_{i=1}^{n_s} \sum_{c=1}^{C} {\rm I}[c = y_i^s] \log p_{i,c}^s,
	\end{aligned}
	\end{eqnarray}
	where $p_{i,c}^s$ is the $c$-th element of category prediction $\bm{p}_i^s$ of a source point cloud $\mathcal{P}_i^s$, and ${\rm I[\cdot]}$ is an indicator function. 
	Supervised learning establishes in feature space $\mathcal{Z}$ a semantic representation $\bm{z}$ that is discriminative among categories on source domain $\mathcal{S}$.
	
	\vspace{0.1cm}\noindent\textbf{Target Domain Self-training with Self-paced Learning --}
	As the ground truth labels of target samples are unavailable, we take a direct strategy of self-training \cite{pseudo_label} that uses pseudo labels to guide the model learning. We have no guarantee in the correctness of the obtained pseudo labels but expect that they are mostly correct. 
	To this end, we employ a self-paced learning in an easy-to-hard learning manner \cite{cbst}, which generates pseudo labels from category predictions at the higher levels of confidence. 
	As a result, the objective of self-paced learning based self-training is depicted as: 
	\begin{align}\label{EqnTrgSelfLearn}
	& \min_{\Phi_\text{fea},\Phi_\text{cls},\widehat{\bm{Y}}^t} \mathcal{L}_\text{cls}^t = - \frac{1}{n_t} \sum_{i=1}^{n_t} \left(  \sum_{c=1}^C \widehat{y}_{i,c}^t \log p_{i,c}^t + \gamma |\widehat{\bm{y}}_i^t|_1 \right) \nonumber \\
	& \quad\: {\rm s.t.} \:\: \widehat{\bm{y}}_i^t \in \{ \{\bm{e} | \bm{e} \in \mathbb{R}^C\} \cup \bm{0} \}, \forall i \in \{1, 2, \cdots, n_t\} \\ 
	& \qquad\quad \gamma > 0, \nonumber
	\end{align}
	where $\widehat{\bm{y}}_i^t$ is the assigned pseudo label vector for a target instance $\mathcal{P}_i^t$, $\widehat{y}_{i,c}^t$ is its $c$-th element, $\widehat{\bm{Y}}^t$ is the set of all pseudo label vectors $\{\widehat{\bm{y}}_i^t\}_{i=1}^{n_t}$, $\bm{e}$ is a one-hot vector, $\bm{0}$ is a $C$-dimensional vector with all zero elements, and $\gamma$ is a hyper-parameter. 
	
	Similar to Eq. (\ref{EqnSrcSupLearn}), the first term in Eq. (\ref{EqnTrgSelfLearn}) aims to maximize the mutual information between the selected input $\mathcal{P}^t$ and its assigned label $\widehat{y}^t$ over the same $\Phi_\text{cls} \circ \Phi_\text{fea}$, giving rise to discriminative features and decision boundaries adapted to the target domain. 
	This indeed makes sense for classification of target samples since the optimal classifiers in individual domains disagree \cite{dirt_t}. 
	In the second term, the negative ${\rm L}_1$ loss is used to avoid degenerate solutions that assign all $\widehat{\bm{y}}^t$ as $\bm{0}$, \ie ignoring all target samples in network training. 
	$\gamma$ controls the number of selected target samples. The larger $\gamma$, the more samples.
	More specifically, the optimization of Eq. (\ref{EqnTrgSelfLearn}) alternates between the following steps.
	\begin{itemize}
		
		\item \emph{Updating pseudo labels --} We first fix the model $\Phi$ and minimize $\mathcal{L}_\text{cls}^t$ in Eq. (\ref{EqnTrgSelfLearn}) over the pseudo label vector set $\widehat{\bm{Y}}^t$. By solving a nonlinear integer programming, we have the optimized solution as follows \cite{cbst}:
		\begin{equation}\label{EqnTrgLblAssign}
		\widehat{y}_{i,c}^t =\! \left\{
		\begin{aligned}
		& 1 \:\: ~~~~~{\rm if} \: c = \arg\max_{c'} p_{i,c'}^t, \: p_{i,c}^t > \exp(-\gamma) \\
		& 0 \:\: ~~~~~{\rm otherwise}.
		\end{aligned}
		\right.
		\end{equation}
		
		\item \emph{Updating the model $\Phi$ --} We then fix all pseudo label vectors $\{\widehat{\bm{y}}_i^t\}_{i=1}^{n_t}$ and minimize $\mathcal{L}_\text{cls}^t$ in Eq. (\ref{EqnTrgSelfLearn}) over the model $\Phi$.
		
	\end{itemize}
	Denote the maximum of $\bm{p}^t$ as the measure of easiness for a target instance $\mathcal{P}^t$, \ie $\max_{c'} p_{c'}^t$. 
	During training, the model iteratively adapts the semantic representation with the more confident target samples.
	With semantic representation increasingly adaptive between the source and target domains, the target samples with less confidence (\ie harder samples) are explored subsequently. 
	Such an easy-to-hard scheme also conforms to the optimization dynamics of supervised learning \cite{CloserLookDN}, which first learns easier examples that better fit patterns.
	
	\vspace{0.1cm}\noindent\textbf{Semantic Representation Learning --}
	By combining source domain supervised learning of Eq. (\ref{EqnSrcSupLearn}) and target domain self-training with self-paced learning of Eq. (\ref{EqnTrgSelfLearn}), we formulate the objective of semantic representation learning:
	\begin{eqnarray}\label{EqnDomSharedSemRepLearn}
	\begin{aligned}
	\min_{\Phi_\text{fea},\Phi_\text{cls},\widehat{\bm{Y}}^t} \mathcal{L}_\text{sem} = \mathcal{L}_\text{cls}^s + \lambda \mathcal{L}_\text{cls}^t,
	\end{aligned}
	\end{eqnarray}
	where $\lambda \in [0,1]$ is a penalty parameter to suppress the noisy signal at the early stage of training. 
	Without explicitly aligning feature across domains, the objective (\ref{EqnDomSharedSemRepLearn}) enforces feature encoder $\Phi_\text{fea}$ to directly output domain-shared semantic representation in $\mathcal{Z}$ with source and target data jointly. 
	Since the two domains share the same label space $\mathcal{Y}$, their samples corresponding to the same categories would ideally be pushed closer to each other in $\mathcal{Z}$, \emph{naturally} achieving feature alignment across domains.

	\subsection{Self-Supervised Geometric Feature Encoding}
	\label{SecGeoEnc}

	The feature output of $\Phi_\text{fea}$ can be ambiguous due to intra-class shape variation, which is made even more challenging for inconsistent distribution of point cloud representation across domains.
	As a result, with a common object classifier $\Phi_\text{cls}$ on the source and target domains, learning domain-invariant geometric feature from point clouds is an alternative solution to improve representation discrimination, which is verified in \cite{PCM_RegRecT}.  
	To this end, we propose to complement the self-training based semantic adaptation with two pretext tasks, \ie the rotation angle prediction and the distortion location prediction, in a self-supervised learning (SSL) fashion, which can model geometric invariance across domains.
	Note that, the proposed self-supervised geometric feature encoding is utilized in feature encoding on both source and target domains, whose supervision signals are generated automatically from the data as other self-supervised learning methods \cite{SSL_PointSeg_PC,PCM_RegRecT}.
	
\begin{figure}[t]
		\centering
		\subfigure[rot $90^\circ$, $90^\circ$]{
			\label{Fig.rotsub.1}
			\begin{minipage}[b]{0.16\textwidth}
				\includegraphics[width=0.9\textwidth]{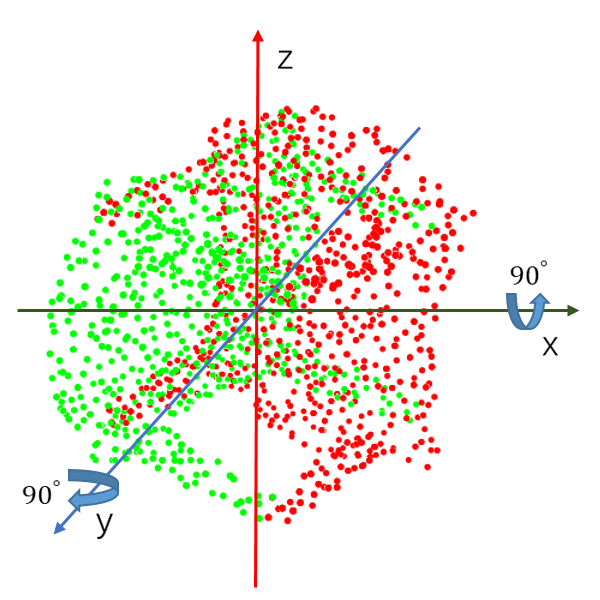} \\
				\includegraphics[width=0.9\textwidth]{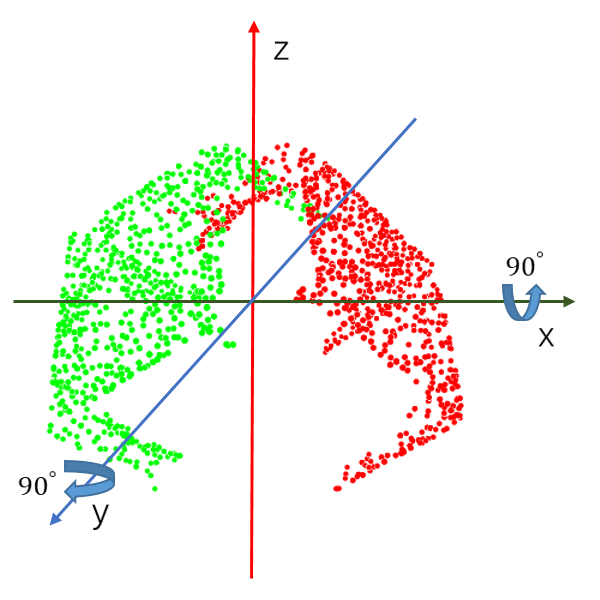}
		\end{minipage}}
		\subfigure[rot $0^\circ$, $180^\circ$]{
			\label{Fig.rotsub.2}
			\begin{minipage}[b]{0.16\textwidth}
				\includegraphics[width=0.9\textwidth]{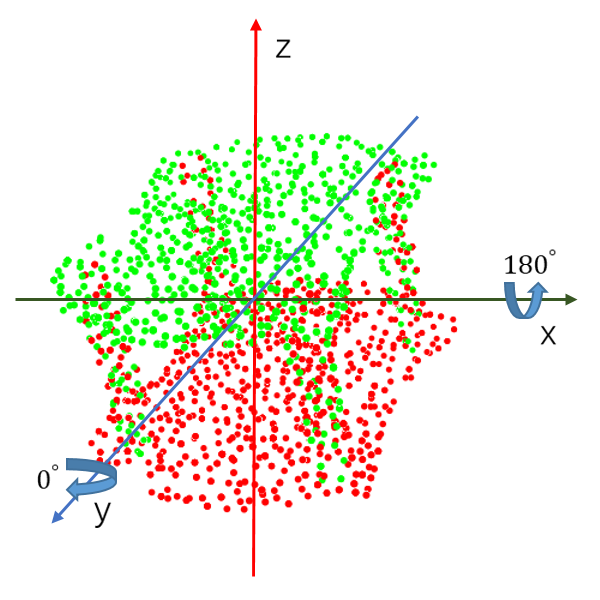} \\
				\includegraphics[width=0.9\textwidth]{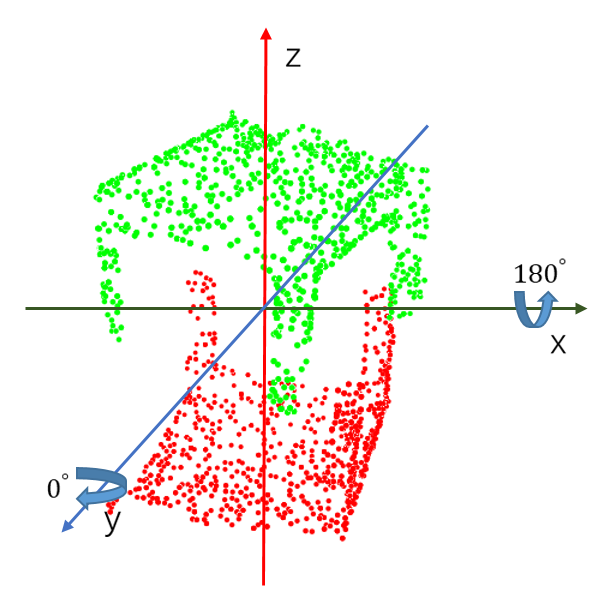}
		\end{minipage}}\vspace{-0.2cm}
		\caption{Illustrative examples of generating two Mixup samples of the Table class, by combining sampled point sub-sets for predicting rotation classes along the $x$-axis and $y$-axis. Two rows correspond to the ModelNet-10 and the ScanNet-10 of the PointDA-10 \cite{PointDAN} respectively.}
		\label{Fig.Rotation}\vspace{-0.6cm}
	\end{figure}

	
	\vspace{0.1cm}\noindent\textbf{Rotation Angle Prediction on Point Cloud Mixup --}
	{Given a point cloud $\mathcal{P}$, we first randomly sample a Mixup coefficient $\alpha \in (0,1)$, which is used to sample two shapes $\mathcal{P}_a \in \mathbb{R}^{\lfloor\alpha\cdot m\rfloor \times 3}$ and $\mathcal{P}_b \in \mathbb{R}^{\lfloor(1-\alpha)\cdot m\rfloor \times 3}$ from $\mathcal{P}$ respectively using farthest point sampling (as in \cite{Pointnet}), where $\lfloor\alpha\cdot m\rfloor$ and $\lfloor(1-\alpha)\cdot m\rfloor$ are the number of sampled points in $\mathcal{P}_a$ and $\mathcal{P}_b$ and $\lfloor \cdot \rfloor$ denote the floor function to output the integers. 
Finally, we form a new point cloud mixup $\widetilde{\mathcal{P}}$ by clockwise rotating the $\mathcal{P}_a$ along the $x$-axis and clockwise rotating the $\mathcal{P}_b$ along the $y$-axis by two randomly-selected angles within $\{0^\circ, 90^\circ, 180^\circ, 270^\circ\}$ respectively.	 
}
To implement an auxiliary task for rotation angle prediction of point cloud mixup, we additionally stack a {Mixup} rotation classifier $\Phi_\text{rot}: \mathcal{Z} \rightarrow [0,1]^R$ on top of the feature extractor $\Phi_\text{fea}(\cdot)$, where $R$ is the number of rotation angle classes. 
Two illustrative examples are given in Figure \ref{Fig.Rotation}.\footnote{Note that, following \cite{PCM_RegRecT,Pointnet} point clouds in our paper are tolerant for arbitrary rotations along the $z$-axis.}	
Following other mixup operation in \cite{mixup}, the label mixup as $\alpha\widetilde{y}_a \cup (1-\alpha)\widetilde{y}_b$ is also generated to form a training pair with corresponding point cloud mixup $\widetilde{\mathcal{P}}\in \mathbb{R}^{m \times 3}$,
where $\widetilde{y}_a \in \{1,2,3,4\} $ and $\widetilde{y}_b \in \{5,6,7,8 \} $ denote the rotation class labels of $\mathcal{P}_a$ and $\mathcal{P}_b$.
	The rotation classifier $\Phi_\text{rot} \circ \Phi_\text{fea}(\widetilde{\mathcal{P}})$ takes as input the rotated point cloud mixup $\{\widetilde{\mathcal{P}}_i\}_{i=1}^{n_s+n_t}$ from both the source and target domains. 
	Given rotation labels $\{\widetilde{y}_{a,i},\widetilde{y}_{b,i} \}_{i=1}^{n_s+n_t}$, we have the following objective for rotation angle classification: 
	\begin{eqnarray}\label{EqnRotPred}
	\begin{aligned}
	\min_{\Phi_\text{fea},\Phi_\text{rot}} \mathcal{L}_\text{rot} = - \frac{1}{n_s+n_t} \sum_{i=1}^{n_s+n_t} \sum_{r=1}^{R}\Big( \alpha{\rm I}[r=\widetilde{y}_{a,i}] \log \widetilde{p}_{i,r} \\ +  (1-\alpha){\rm I}[r=\widetilde{y}_{b,i}] \log \widetilde{p}_{i,r}\Big), 
	\end{aligned}
	\end{eqnarray}
	{where $R=8$ and $\widetilde{p}_{i,r}$ is the $r$-th element of the predicted rotation probability vectors $\widetilde{\bm{p}}_{i} = \Phi_\text{rot} \circ \Phi_\text{fea}(\widetilde{\mathcal{P}}_{i})$.} 
	Optimizing the objective (\ref{EqnRotPred}) enables the model to perceive global and topological configuration of local shape primitives in 3D space as \cite{SSL_RotPred_PC}.
	
\begin{figure}[t]
		\centering
		\subfigure[loc 1]{
			\label{Fig.locsub.1}
			\begin{minipage}[b]{0.16\textwidth}
				\includegraphics[width=1.0\textwidth]{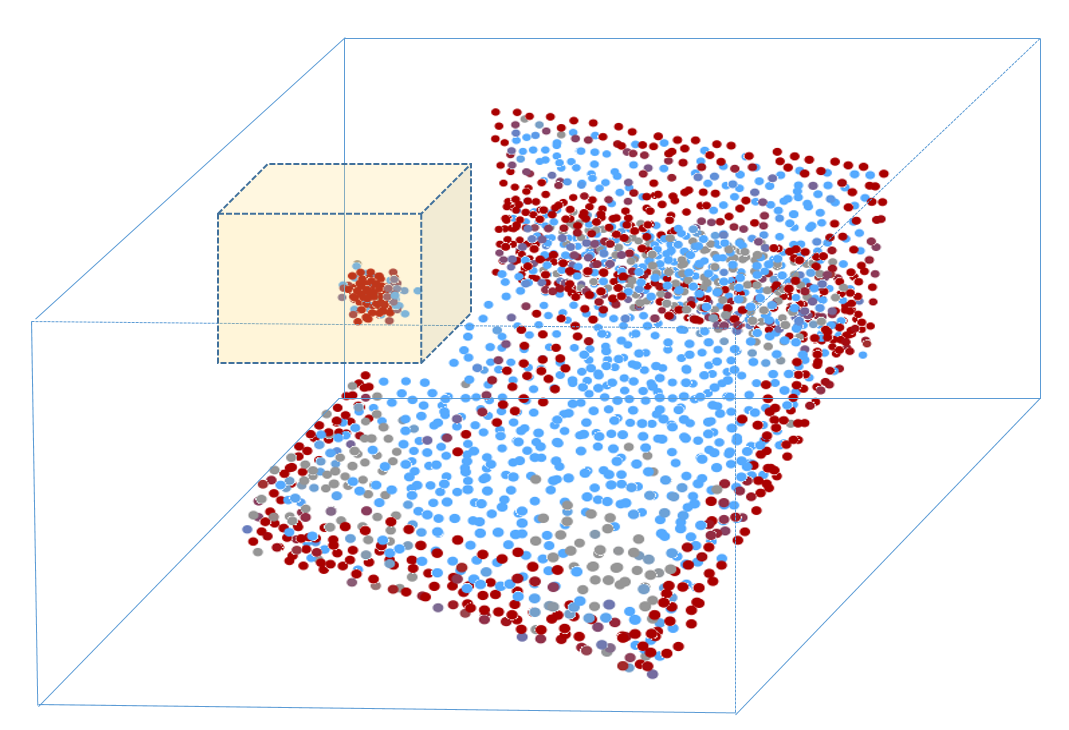} \\
				\includegraphics[width=1.0\textwidth]{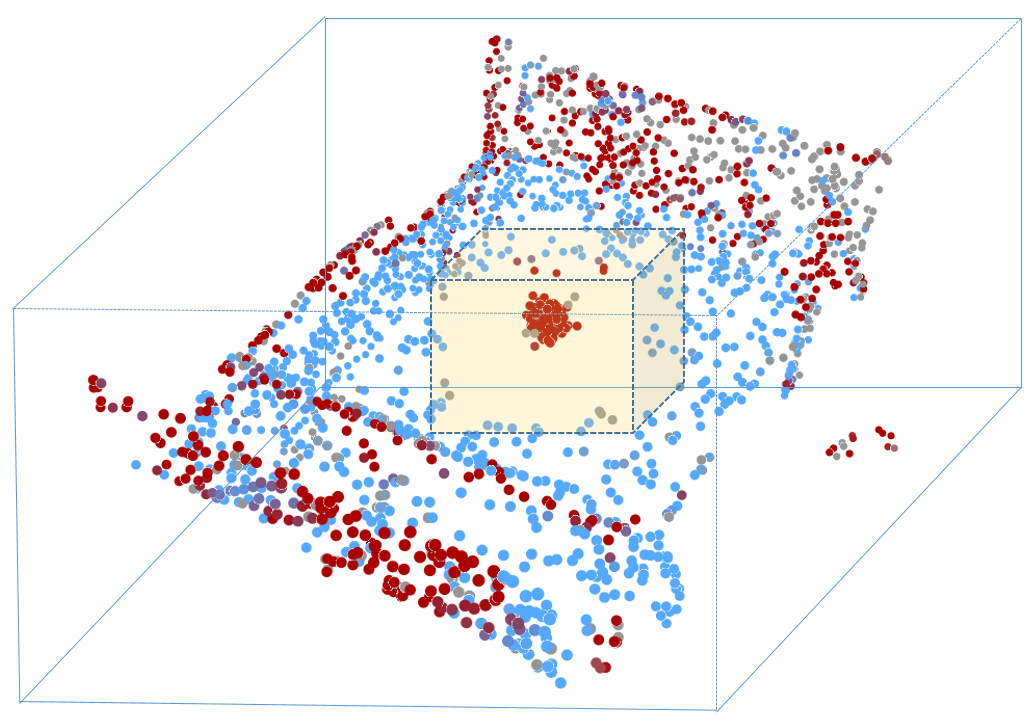}
		\end{minipage}}
		\subfigure[loc 2]{
			\label{Fig.locsub.2}
			\begin{minipage}[b]{0.16\textwidth}
				\includegraphics[width=1.0\textwidth]{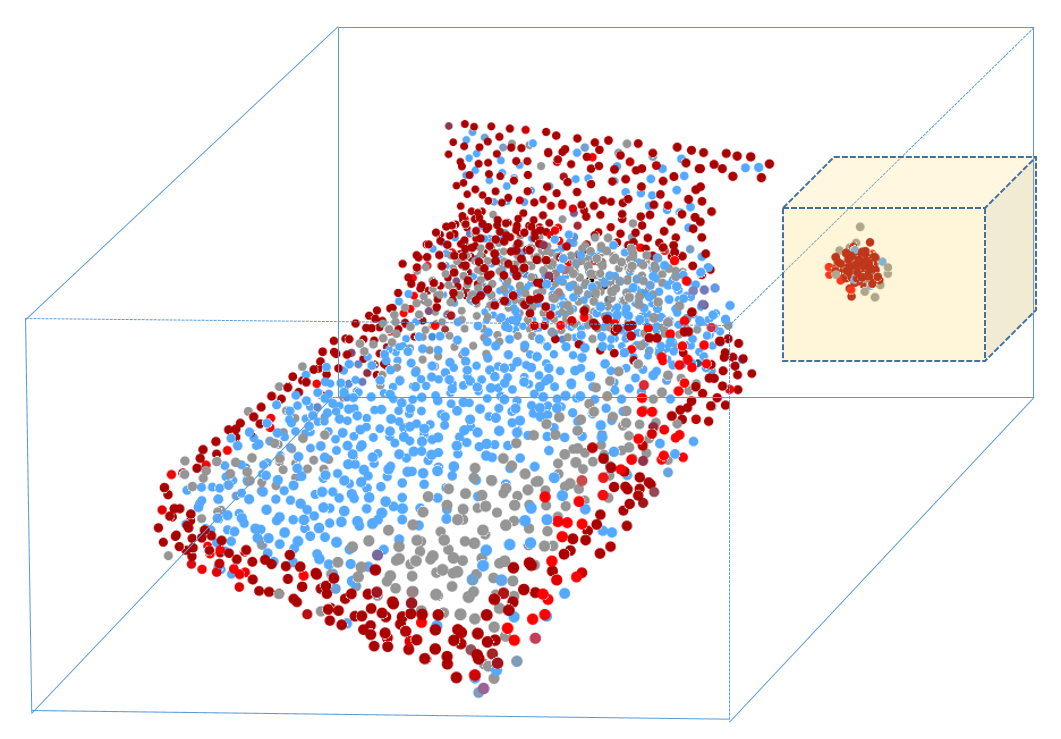} \\
				\includegraphics[width=1.0\textwidth]{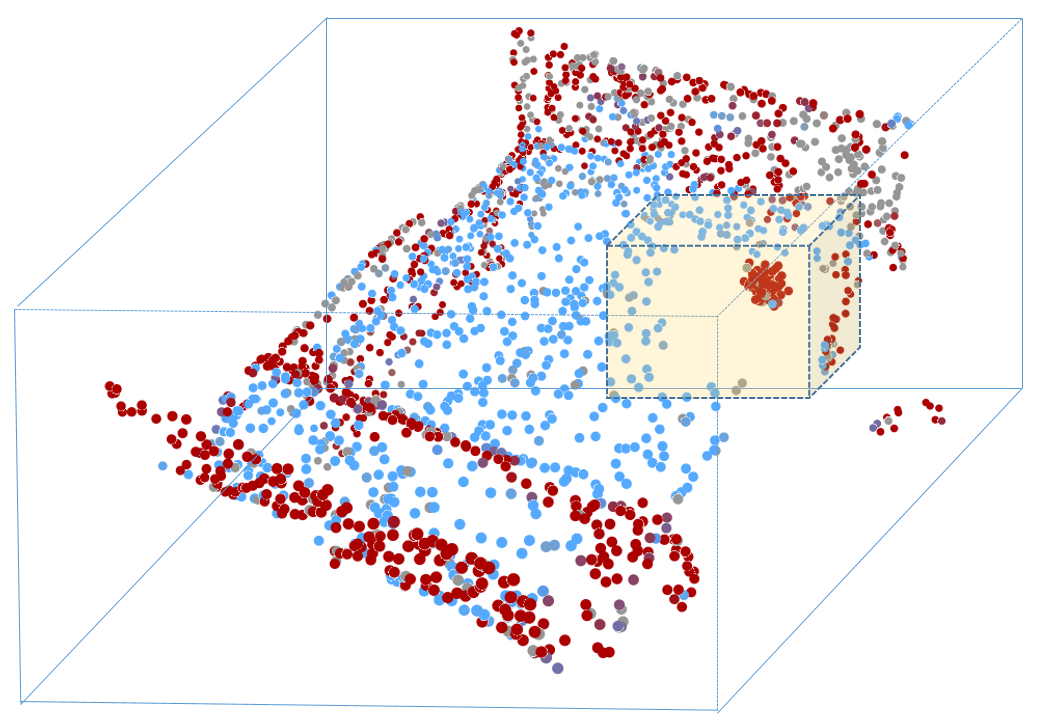}
		\end{minipage}}\vspace{-0.2cm}
		\caption{Illustration of bed examples with different distortion locations {and corresponding curvature-based target codes for classification.} Two rows correspond to the ModelNet-10 and the ScanNet-10 domains of the PointDA-10 \cite{PointDAN} respectively. }
		\label{Fig.Location}\vspace{-0.6cm}
	\end{figure}

	\vspace{0.1cm}\noindent\textbf{Curvature-Aware Distortion Localization --}
	To incorporate local geometries into feature representation, the state-of-the-art methods \cite{SSL_UDA_Images,SSL_PointSeg_PC,PCM_RegRecT} have explored the pretext tasks \wrt location or distortion, \eg reconstructing a point cloud with randomly displaced or distorted parts. 
	Inspired by their success, this paper proposes a simple yet effective pretext task, \ie~{predicting the location distorted point set with explicitly incorporating geometric property -- curvature.
	Intuitively, the higher curvature, the richer geometric information preserve \cite{cad}.
	 To this end, we first obtain curvature of each point by direct physical computation based on principle component analysis (PCA) within a local region, \eg seeking an optimal plane that best fitting the central point and its k-nearest neighbours.}
	We then voxelize the point cloud $\mathcal{P}$ into $k^3$ voxels, from which we randomly select one at equal probability and replace all points within such a voxel with an equal number of points sampled from an isotropic Gaussian distribution, where the mean is the center of the sampled voxel and the standard deviation is typically small. 
	Instead of using one-hot target coding, a soft target code revealing local geometries, \eg pointwise geometric property such as curvature, can enforce the network to focus on regions with higher curvature in a cost-sensitive learning manner.  
    In details, the curvature cost is the ratio of the curvature sum of any point subset to be distorted to those of the whole point cloud.
	By taking these steps, the distorted point clouds $\{\overline{\mathcal{P}}_i\}_{i=1}^{n_s+n_t}$ for both domains are produced, as shown in Figure \ref{Fig.Location}.
	{{Such a distortion localization with a curvature-sensitive label can be formulated into a classification problem.} }
    Specifically speaking, we stack a location classifier $\Phi_\text{loc}: \mathcal{Z} \rightarrow [0, 1]^L$ on top of $\Phi_\text{fea}(\cdot)$ for discovering one distorted geometric cell from the whole point cloud, where $L = k^3$ is the number of voxels to cover all the shape surface. 
	Given the indexes of the chosen cells as location labels $\{\overline{y}_i \in \{1,2,\cdots,L\}\}_{i=1}^{n_s+n_t}$ {and the corresponding curvature cost $\{\overline{c}_i \in R \}_{i=1}^{n_s+n_t}$}, we train the location prediction model $\Phi_\text{loc} \circ \Phi_\text{fea}$ by the following objective:
	\begin{eqnarray}
	\begin{aligned}
	\min_{\Phi_\text{fea},\Phi_\text{loc}} \mathcal{L}_\text{loc} = - \frac{1}{n_s+n_t} \sum_{i=1}^{n_s+n_t} \overline{c}_i\ \sum_{l=1}^L {\rm I}[l=\overline{y}_i] \log \overline{p}_{i,l} 
	\end{aligned}\label{EqnLocPred}
	\end{eqnarray}
	where 
	$\overline{p}_{i,l}$ is the $l$-th element of the predicted location probability vector $\overline{\bm{p}}_{i} = \Phi_\text{loc}\circ \Phi_\text{fea}(\overline{\mathcal{P}}_i)$. Intuitively, the objective (\ref{EqnLocPred}) can capture the geometric information from local distributions of spatial points in a point cloud via infering where the distorted points are.
	
	\vspace{0.1cm}\noindent\textbf{Geometric Feature Encoding --}
	By combining the two self-supervised learning objectives (\ref{EqnRotPred}) and (\ref{EqnLocPred}), we form the objective of geometry encoding as:
	\begin{eqnarray}\label{EqnGeoInfoEnc}
	\begin{aligned}
	\min_{\Phi_\text{fea},\Phi_\text{rot},\Phi_\text{loc}} \mathcal{L}_\text{geo} = \mathcal{L}_\text{rot} + \mathcal{L}_\text{loc}. 
	\end{aligned}
	\end{eqnarray}
	Since feature encoding function $\Phi_\text{fea}$ is shared across semantic feature adaption and geometric feature learning, the objective (\ref{EqnGeoInfoEnc}) forces  $\Phi_\text{fea}$ to learn expressive features capturing global and local geometric invariance across domains via a proxy loss of the two pretext tasks.
	In other words, by learning global rotation and local deformation of point clouds $\mathcal{P}$ in both $\mathcal{S}$ and $\mathcal{T}$ domains, the domain discrepancy is further reduced in view of construction of common geometric feature space using both source and target data.
	Specifically, in self-supervised learning tasks, both source and target data will be assigned an automatically generated rotation/distortion index label, and geometric features jointly learned from both source and target data can thus be more robust against point distribution variations in different domains.

	\subsection{Overall Training and Inference}
	The overall training objective integrates semantic representation learning (\ref{EqnDomSharedSemRepLearn}) and geometry-aware regularization (\ref{EqnGeoInfoEnc}), leading to a unified framework of \emph{Geometry-Aware Self-Training (GAST)} for UDA, as follows:
	\begin{eqnarray}\label{EqnOverallObj}
	\begin{aligned}
	\min_{\Phi_\text{fea},\Phi_\text{cls},\widehat{\bm{Y}},\Phi_\text{rot},\Phi_\text{loc}} \mathcal{L}_\text{GAST} = \mathcal{L}_\text{sem} + \beta \mathcal{L}_\text{geo},
	\end{aligned}
	\end{eqnarray}
	where $\beta$ is a trade-off hyper-parameter and all model parameters of the proposed GAST (cf. Figure \ref{Fig.Architecture}) are simultaneously learned in an end-to-end manner. 
	Once trained, our model can be simply deployed as a conventional classification model by discarding the classifiers of rotation and distortion location. 
	During testing, we infer the category label for any target test point cloud $\mathcal{P}$ as $\arg\max_c p_c$, where $p_c$ is the $c$-th element of the predicted category probability vector $\bm{p}=\Phi_\text{cls} \circ \Phi_\text{fea}(\mathcal{P})$.
	
	\section{Experiments}
	
	\begin{table*}[ht]
		\centering
		\resizebox{0.92\linewidth}{!}{
			\begin{tabular}{lccc||ccccccc}
				\hline
				& LocCls & RotCls & SPST & M$\rightarrow$S & M$\rightarrow$S* & S$\rightarrow$M & S$\rightarrow$S* & S*$\rightarrow$M & S*$\rightarrow$S & Avg.\\
				\hline\hline
				Supervised & & & & 93.9 $\pm$ 0.2 & 78.4 $\pm$ 0.6 & 96.2 $\pm$ 0.1 & 78.4 $\pm$ 0.6 & 96.2 $\pm$ 0.1 & 93.9 $\pm$ 0.2 & 89.5 $\pm$ 0.3 \\
				w/o Adapt & & & & 83.3 $\pm$ 0.7 & 43.8 $\pm$ 2.3 & 75.5 $\pm$ 1.8 & 42.5 $\pm$ 1.4 & 63.8 $\pm$ 3.9 & 64.2 $\pm$ 0.8 & 62.2 $\pm$ 1.8 \\
				\hline
				DANN \cite{dann} & & & & 74.8 $\pm$ 2.8 & 42.1 $\pm$ 0.6 & 57.5 $\pm$ 0.4 & 50.9 $\pm$ 1.0 & 43.7 $\pm$ 2.9 & 71.6 $\pm$ 1.0 & 56.8 $\pm$ 1.5 \\
				PointDAN \cite{PointDAN} & & & & 83.9 $\pm$ 0.3 & 44.8 $\pm$ 1.4 & 63.3 $\pm$ 1.1 & 45.7 $\pm$ 0.7 & 43.6 $\pm$ 2.0 & 56.4 $\pm$ 1.5 & 56.3 $\pm$ 1.2 \\
				RS \cite{SSL_PointSeg_PC} & & & & 79.9 $\pm$ 0.8 & 46.7 $\pm$ 4.8 & 75.2 $\pm$ 2.0 & 51.4 $\pm$ 3.9 & 71.8 $\pm$ 2.3 & 71.2 $\pm$ 2.8 & 66.0 $\pm$ 1.6 \\
				DefRec + PCM \cite{PCM_RegRecT} & & & & 81.7 $\pm$ 0.6 & 51.8 $\pm$ 0.3 & 78.6 $\pm$ 0.7 & 54.5 $\pm$ 0.3 & 73.7 $\pm$ 1.6 & 71.1 $\pm$ 1.4 & 68.6 $\pm$ 0.8 \\
				\hline
				\multirow{5}{*}{\textbf{GAST}} 
				& \checkmark & & & 78.6 $\pm$ 0.3 & 52.3 $\pm$ 0.2 & 75.0 $\pm$ 0.2 & 51.4 $\pm$ 0.3 & 69.3 $\pm$ 0.2 & 63.6 $\pm$ 0.2 & 65.1 $\pm$ 0.2 \\
				& & \checkmark & & 84.3 $\pm$ 0.2 & 46.2 $\pm$ 0.3 & 69.8 $\pm$ 0.6 & 49.2 $\pm$ 0.3 & 66.6 $\pm$ 0.5 & 66.1 $\pm$ 0.2 & 63.7 $\pm$ 0.4 \\
				& & & \checkmark & 84.4 $\pm$ 0.4 & 45.9 $\pm$ 0.5 & 80.5 $\pm$ 0.3 & 48.7 $\pm$ 0.4 & 64.8 $\pm$ 0.3 & 70.4 $\pm$ 0.3 & 65.8 $\pm$ 0.4 \\
				& \checkmark & \checkmark & & 83.9 $\pm$ 0.2 & 56.7 $\pm$ 0.3 & 76.4 $\pm$ 0.2 & 55.0 $\pm$ 0.2 & 73.4 $\pm$ 0.3 & 72.2 $\pm$ 0.2 & 69.5 $\pm$ 0.2 \\
				& \checkmark & \checkmark & \checkmark & \textbf{84.8} $\pm$ 0.1 & \textbf{59.8} $\pm$ 0.2 & \textbf{80.8} $\pm$ 0.6 & \textbf{56.7} $\pm$ 0.2 & \textbf{81.1} $\pm$ 0.8 & \textbf{74.9} $\pm$ 0.5 & \textbf{73.0} $\pm$ 0.4 \\
				\hline
			\end{tabular}
		}
		\caption{Comparative evaluation in classification accuracy (\%) averaged over 3 seeds ($\pm$ SEM) on the PointDA-10 dataset.} 
		\label{table:PointDA10_comparison}\vspace{-0.2cm}
	\end{table*}
	
	\begin{table*}[ht]
		\small
		\centering
		\resizebox{0.92\linewidth}{!}{
			\begin{tabular}{l  ccc||ccccccccccc}
				\hline
				& LocCls & RotCls & SPST & Bathtub & Bed & Bookshelf & Cabinet & Chair & Lamp & Monitor & Plant & Sofa & Table & Avg.\\
				\hline\hline
				Supervised & & & & 76.9 & 58.8 & 55.5 & 73.2 & 92.5 & 63.4 & 70.5 & 72.0 & 56.0 & 85.0 & 70.4\\
				w/o Adapt & & & & 61.5 & 31.8 & 32.9 & 0 & 49.8 & 36.6 & 54.1 & {96.0} & 30.6 & 47.5 & 44.1\\
				\hline
				DANN \cite{dann} & & & & 34.6 & 38.8 & 34.2 & 2.7 & 59.4 & 12.2 & 49.2 & 84.0 & \textbf{53.0} & 57.8 & 42.6 \\
				PointDAN \cite{PointDAN} & & & & 34.6 & 36.5 & 35.6 & \textbf{3.4} & 61.2 & 29.3 & 37.7 & 76.0 & 44.8 & 45.5 & 40.4 \\
				DefRec + PCM \cite{PCM_RegRecT} & & & & \textbf{65.4} & \textbf{49.4} & 49.3 & 1.3 & 61.4 & 41.4 & 55.7 & 88.0 & 42.5 & 60.8 & 51.5\\
				\hline
				\multirow{5}{*}{\textbf{GAST}}
				& \checkmark & & & 61.5 & 31.8 & \textbf{51.4} & 2.0 & 61.8 & 34.1 & 32.8 & 76.0 & 41.0 & 65.1 & 45.8\\
				& & \checkmark & & 53.8 & 28.2 & 37.7 & 2.0 & 54.9 & 7.3 & 63.9 & 84.0 & 40.3 & 62.5 & 43.5\\
				& & & \checkmark & 57.7 & 35.3 & 45.2 & 3.3 & 54.3 & 34.1 & 49.2 & 76.0 & 51.5 & 50.2 & 45.7\\
				& \checkmark & \checkmark & & 61.5 & 44.7 & 41.8 & \textbf{3.4} & 70.2 & 39.0 & 68.9 & 88.0 & 38.1 & 66.8 & 52.2\\
				& \checkmark & \checkmark & \checkmark & 57.7 & 38.8 & 35.6 & 2.0 & \textbf{74.3} & \textbf{43.9} & \textbf{77.0} & \textbf{96.0} & 45.5 & \textbf{74.1} & \textbf{54.5}\\
				\hline
			\end{tabular}
		}
		\caption{Evaluation of class-wise classification accuracy (\%) on the ModelNet-10 to the ScanNet-10 (M$\rightarrow$S*).} 
		\label{table:classwise_comparision}\vspace{-0.6cm}
	\end{table*}
	
	
	\subsection{Dataset and Settings}
	
	\vspace{0.1cm}\noindent\textbf{Dataset --}
	The PointDA-10 \cite{PointDAN} collects object point clouds of 10 shared classes from the  ModelNet40 \cite{Modelnet}, the ShapeNet \cite{Shapenet} and the ScanNet \cite{Scannet}, leading to the following three distinct domains. 
	\textbf{(1)} the ModelNet-10 (\textbf{M}) consists of 4,183 training and 856 testing point clouds by sampling 2,048 points from the surface of clean 3D CAD models by following the method \cite{Pointnet++}. 
	\textbf{(2)} ShapeNet-10 (\textbf{S}) includes 17,378 training and 2,492 testing point clouds uniformly sampled on the surface of ShapeNet objects, each one also containing 2,048 points. 
	Note that, the ShapeNet-10 is more heterogeneous than the ModelNet-10 since the ShapeNet has more object instances, among which a larger structure variance exists. 
	\textbf{(3)} ScanNet-10 (\textbf{S*}) comprises 6,110 training and 1,769 testing samples
	which collect 2,048 points from partially visible object point clouds of the ScanNet, within manually annotated bounding boxes. 
	As the ScanNet contains point clouds of scanned and reconstructed real-world scenes, point clouds are usually incomplete in view of occlusion with contextual objects in the scenes in addition to realistic sensor noises. 
	We follow the data preparation and data settings used in \cite{PCM_RegRecT}.
	Specifically, all object point clouds in all domains (\ie datasets) are aligned along the $x$ and $y$ axes, only tolerant for arbitrary rotations along the $z$ axis.
	Moreover, a point subset containing 1,024 points are down-sampled from the original 2,048 point clouds provided by the PointDA-10 and is normalized within a unit ball with random jittering as \cite{Pointnet}, which is adopted in all the methods for a fair comparison.
	A typical 80\%/20\%  data split for training and testing on both source and target domains is employed \cite{PCM_RegRecT}.

	
	\vspace{-0.1cm}\noindent\textbf{Comparative Methods --} 
	We compare our proposed GAST with a serial of representative UDA methods on image classification and the state-of-the-art point-based DA methods including Domain Adversarial Neural Network (\textbf{DANN}) \cite{dann}, Point Domain Adaptation Network (\textbf{PointDAN}) \cite{PointDAN}, Reconstruction Space Network (\textbf{RS}) \cite{SSL_PointSeg_PC}, and Deformation Reconstruction Network with Point Cloud Mixup (\textbf{DefRec + PCM}) \cite{PCM_RegRecT}. 
	The \textbf{Supervised} method, that trains the same backbone classifier $\Phi_\text{cls} \circ \Phi_\text{fea}$ with labeled target data only, and the \textbf{w/o Adapt} method that trains the identical backbone net with only labeled source samples, are also evaluated as references of the upper and lower performance bounds, respectively. 
	All comparative methods take the same training protocol and the best models are selected according to source-validation based early stopping.
	
	\vspace{0.1cm}\noindent\textbf{Implementation Details --}
	For our GAST, we adopt the DGCNN \cite{DGCNN} as backbone of the \textit{Feature Encoder} $\Phi_\text{fea}$, while the \textit{Category Classifier} $\Phi_\text{cls}$ is based on a multi-layer perceptron (MLP) with three fully connected (FC) layers (\ie $\{512,256, 10\}$) in view of 10 semantic classes in the PointDA-10. 
	For self-supervised rotation and distorted part classifiers, the GAST respectively employs two two-layer MLPs (\ie $\{512, 4\}$) in view of Mixup rotation angle classification mentioned in Sec. \ref{SecGeoEnc}, and a three-layer MLP (\ie $\{512,256, 27\}$) where the whole object surface is voxelized into $3^3$ cells. 
	The hyper-parameters of $\gamma$ and $\beta$ are empirically set to 0.05 and 1 respectively.
	During training, on the selected target samples with pseudo labels, we follow \cite{noisy_student} to augment data with random rotation along the $z$-axis. 
	The Adam optimizer \cite{adam} is utilized with the initial learning rate 0.001, weight decay 0.00005 and an epoch-wise cosine annealing learning rate scheduler. 
	In total, we train all the methods for 150 epochs with batch size 16 on an NVIDIA GTX-1080 Ti GPU and perform three trials of different random seeds.



	\begin{figure*}[t]
		\centering
		\subfigure[w/o Adapt: M $\rightarrow$ S*]{
			\label{Fig.cofmt_unsu_m2r.1}
			\begin{minipage}[b]{0.24\textwidth}
				\includegraphics[width=1\textwidth]{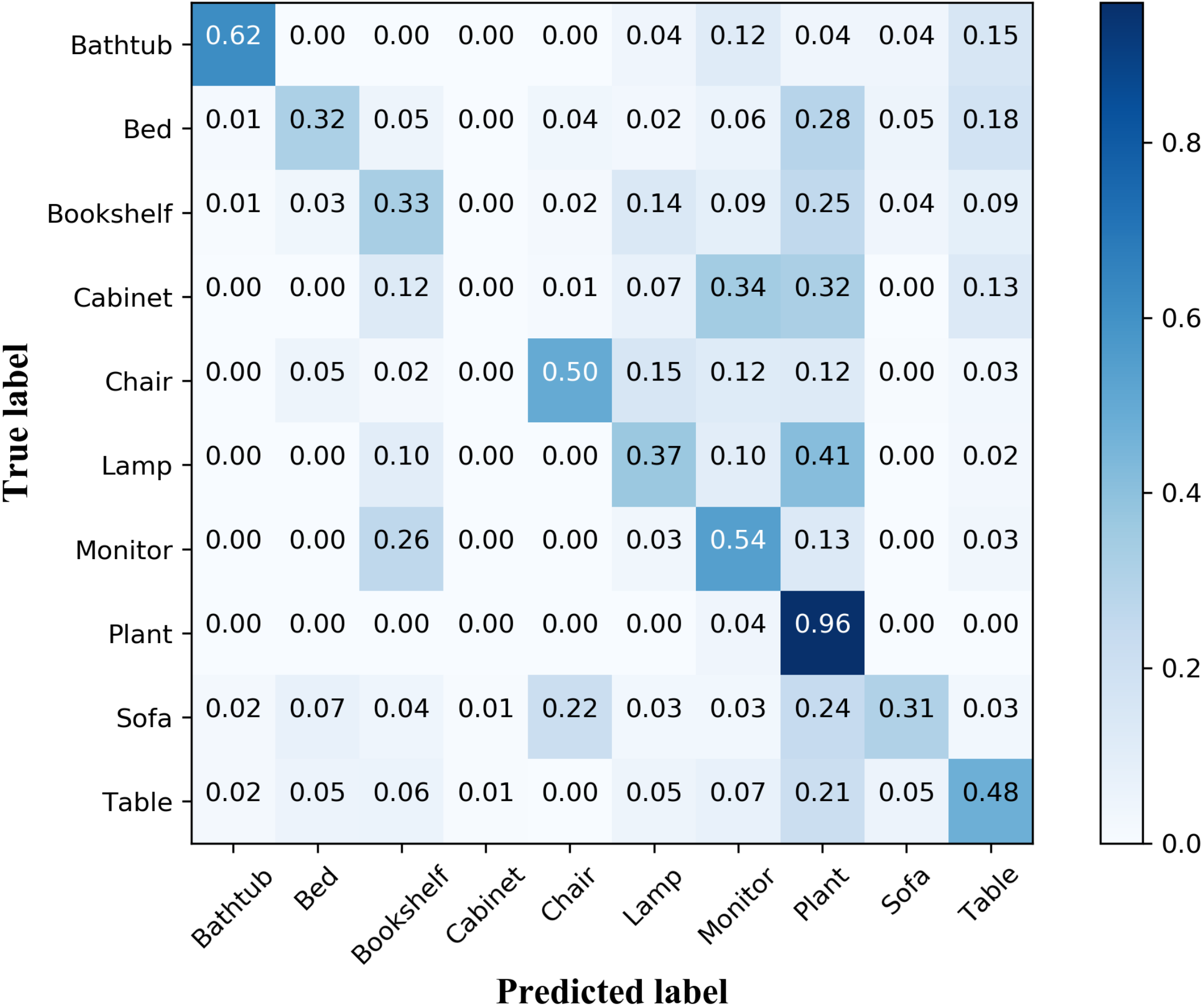}
		\end{minipage}}
		\subfigure[GAST: M $\rightarrow$ S*]{
			\label{Fig.cofmt_m2r.2}
			\begin{minipage}[b]{0.24\textwidth}
				\includegraphics[width=1\textwidth]{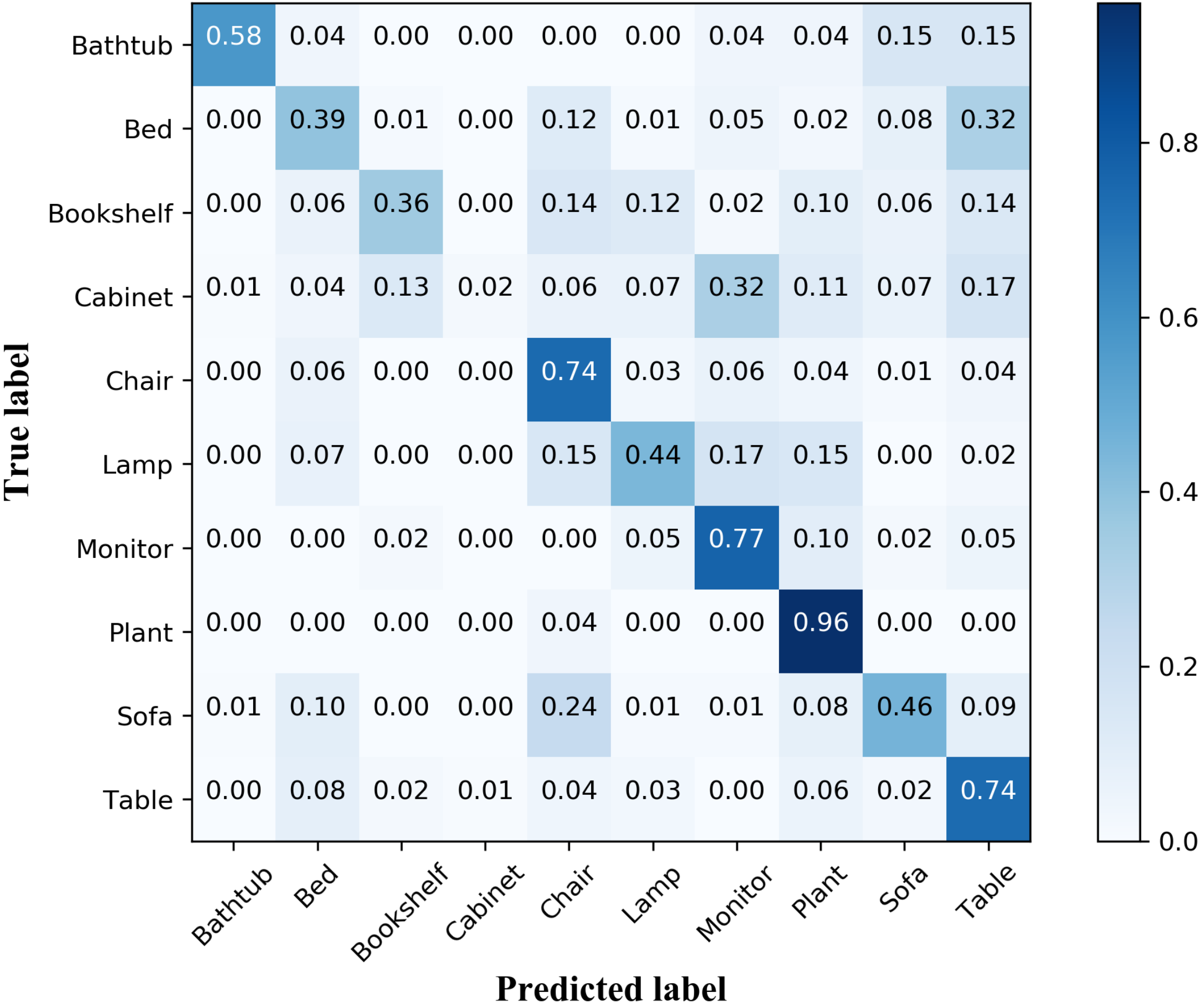}
		\end{minipage}}
		\subfigure[w/o Adapt: S $\rightarrow$ S*]{
			\label{Fig.cofmt_unsu_s2r.3}
			\begin{minipage}[b]{0.24\textwidth}
				\includegraphics[width=1\textwidth]{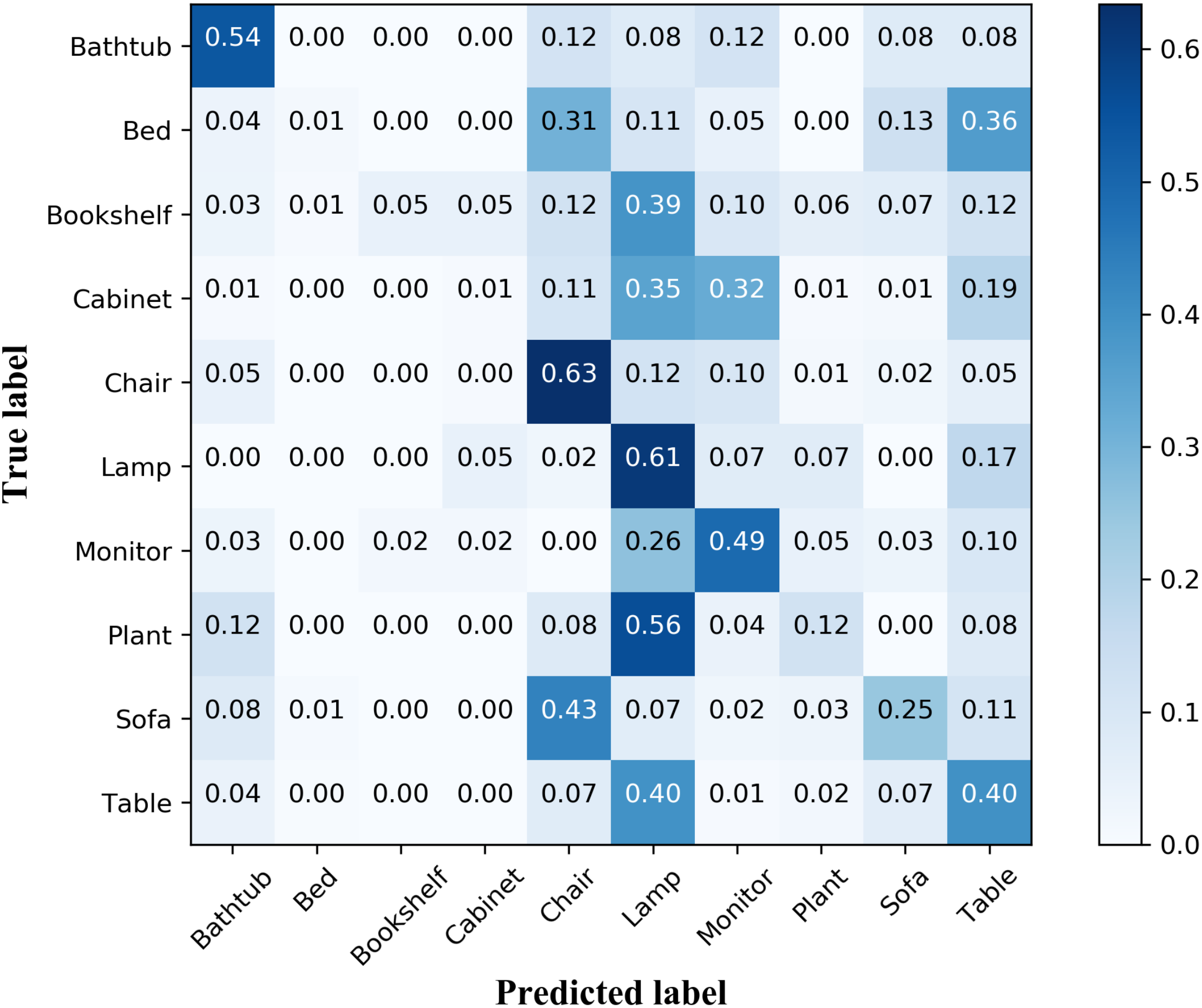}
		\end{minipage}}
		\subfigure[GAST: S $\rightarrow$ S*]{
			\label{Fig.cofmt_s2r.4}
			\begin{minipage}[b]{0.24\textwidth}
				\includegraphics[width=1\textwidth]{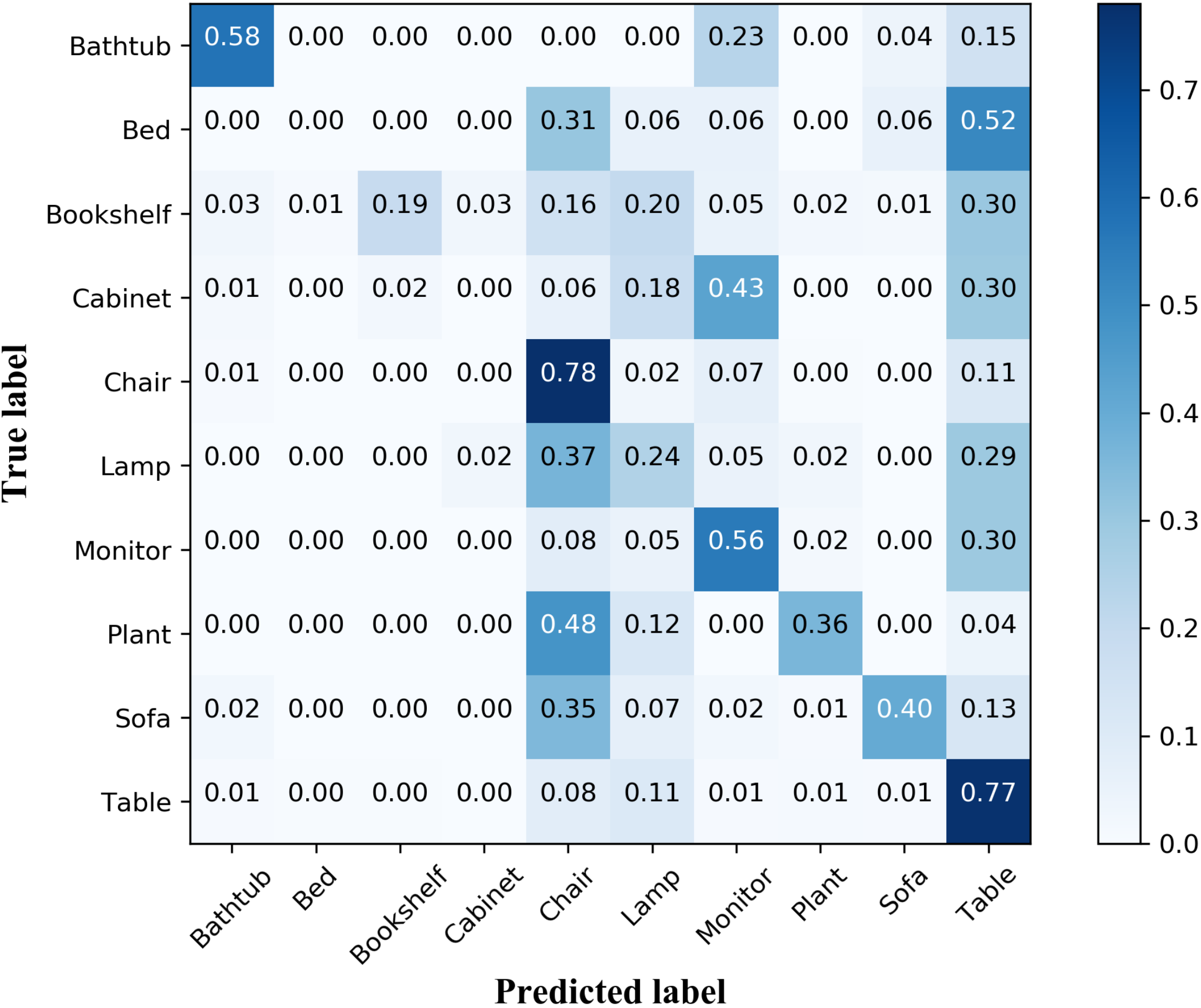}
		\end{minipage}}\vspace{-0.4cm}
		\caption{Confusion matrices of classifying testing samples on target domain.} 
		\label{Fig.confusion_matrix}\vspace{-0.7cm}
	\end{figure*}
	
	\begin{figure}[t]
		\centering
		\subfigure[w/o Adapt: S* $\rightarrow$ M]{
			\label{Fig.embed_unsu_r2m.1}
			\begin{minipage}[b]{0.19\textwidth}
				\includegraphics[width=0.9\textwidth]{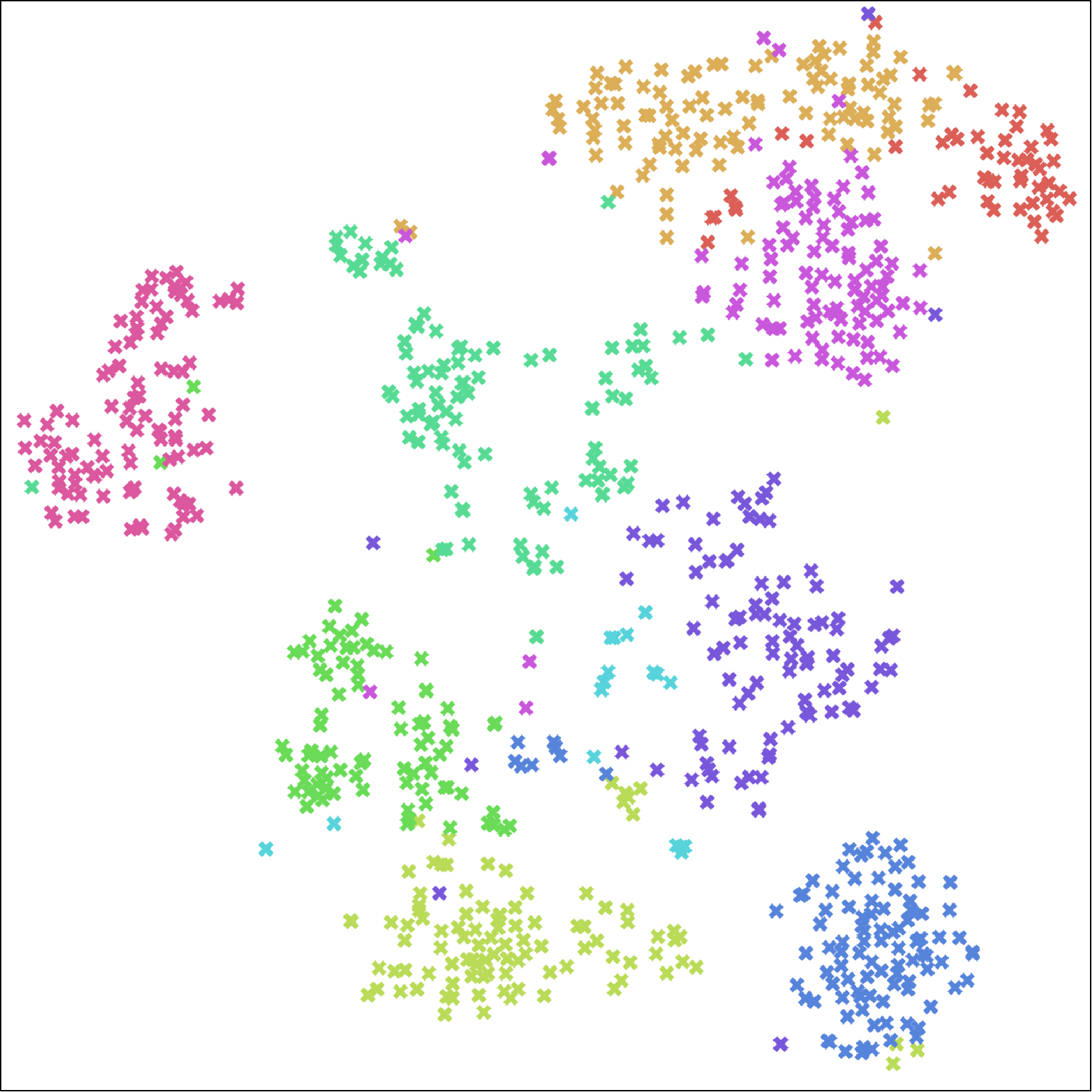}
		\end{minipage}}
		\subfigure[GAST: S* $\rightarrow$ M]{
			\label{Fig.embed_r2m.2}
			\begin{minipage}[b]{0.19\textwidth}
				\includegraphics[width=0.9\textwidth]{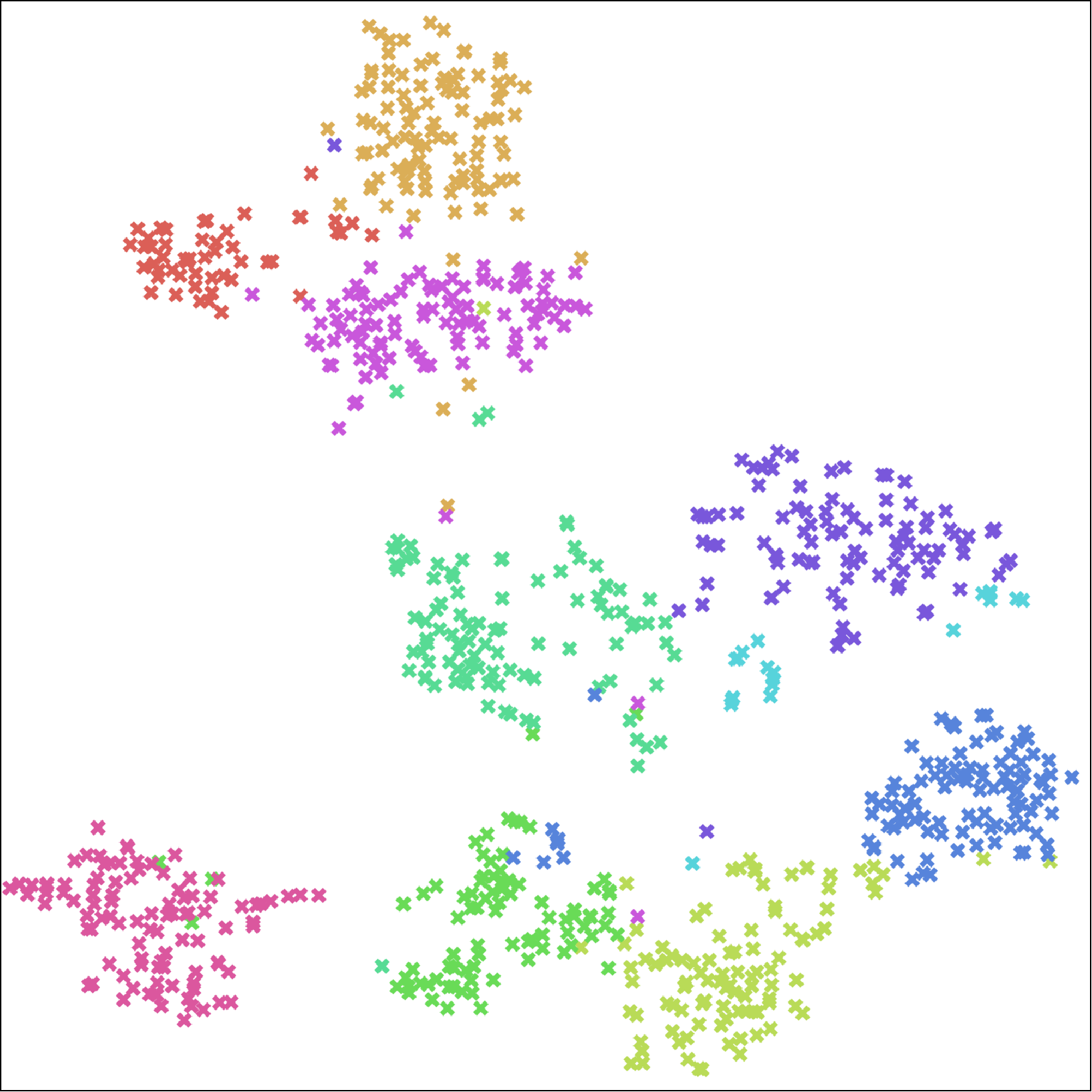}
		\end{minipage}}\vspace{-0.2cm}
		\caption{The t-SNE visualization of feature distribution on the target domain. Colors indicate different classes.}
		\label{Fig.embed_feature}\vspace{-0.7cm}
	\end{figure}

	\subsection{Results}
	
	\vspace{0.1cm}\noindent\textbf{Comparison with the State-of-the-art Methods --}
	Table \ref{table:PointDA10_comparison} presents comparisons between our proposed GAST and other competing methods on the PointDA-10 dataset. 
	Evidently, the GAST\footnote{The GAST in this paragraph indicates the variant with all components.} outperforms all comparative domain adaptation methods with a significant margin, with improving the average accuracy by 4.4\% and  16.7\%  over the state-of-the-art DefRec + PCM \cite{PCM_RegRecT} and the PointDAN \cite{PointDAN} respectively.
	It is noteworthy that for the challenging yet realistically significant synthetic-to-real tasks of $\textbf{M $\rightarrow$ S*}$ and $\textbf{S $\rightarrow$ S*}$, the GAST acquires a remarkable enhancement over w/o Adapt by 16\% and 14.2\% respectively.
	Visualization of confusion matrices in terms of class-wise classification accuracy achieved by {the w/o Adapt} and our GAST on two synthetic-to-real UDA tasks of M $\rightarrow$ S* and S $\rightarrow$ S* shown in Figure \ref{Fig.confusion_matrix}. 
	As the w/o Adapt baseline and the proposed GAST use the identical DGCNN backbone, significant performance gain (10.8\% on average) of our method over the baseline can only be credited to the design of our geometry-aware self-training method to learn a better semantic representation, which is more adaptive across domains and more discriminative among categories,
	benefiting from integrating self-paced semantic adaptation with self-supervised geometric encoding. 
	More importantly, in comparison with the DefRec + PCM, the proposed GAST achieves superior performance on two synthetic-to-real $\textbf{M $\rightarrow$ S*}$ and $\textbf{S $\rightarrow$ S*}$ tasks with 8.0\% and 2.2\% performance gain respectively.
	Class-wise classification accuracy on the task M $\rightarrow$ S* is also reported in Table \ref{table:classwise_comparision}.
	Compared with existing methods, the proposed GAST achieves the better performance on most of the classes, especially those major classes with much more samples such as Chair and Table classes.
	Results of our method for long-tailed classes such as the Bathtub and Bed classes can be comparable to those of the DefRec + PCM, but our method remains its superiority to the DANN and the PointDAN.
	The performance gap between head and long-tailed classes of our method can be explained by the representation learning demanding sufficient samples to characterize semantics across domains. 

	\vspace{0.1cm}\noindent\textbf{Ablation Studies --} 
	We examine the effects of three key components of our GAST, \ie \textbf{LocCls} (distortion location prediction), \textbf{RotCls} (rotation angle prediction), and \textbf{SPST} (self-paced self-training) respectively. 
	Tables \ref{table:PointDA10_comparison} and \ref{table:classwise_comparision} compare the five GAST variants that contain different combinations of these components: \textbf{(1)} LocCls only, \textbf{(2)} RotCls only, \textbf{(3)} SPST only, \textbf{(4)} RotCls + LocCls, and \textbf{(5)} RotCls + LocCls + SPST. We highlight the main observations below. First, each component has a positive impact and 
	method (5) with all the components achieves the best performance, verifying that all components of our GAST are complementary. 
	Second, self-supervised geometric encoding is effective to handle with distribution shifts of point-based shape representation, whose results (\ie RotCls + LocCls) without the SPST can still outperform the state-of-the-art methods, which can be explained by joint representation learning on source and target data with self-generated labels to capture common geometric patterns across domains.

	
	\vspace{0.1cm}\noindent\textbf{Feature Visualization --} We utilize t-SNE \cite{t-SNE} to visualize the feature distribution on the target domain of the UDA task S* $\rightarrow$ M of the baseline and our GAST in Figure \ref{Fig.embed_feature}. 
	In view of an imbalanced data distribution, features of the head classes (\eg the yellow and green ones) with more samples are emphasized during representation learning, and thus can be more discriminative than those of the baseline. 
	
	\section{Conclusion}
	This work aims to learn a domain-shared representation of semantic categories on point clouds via a novel Geometry-Aware Self-Training (GAST) method.
	Experiments on the PointDA-10 benchmark can verify the effectiveness of key components in our scheme, achieving the new state-of-the-art performance.
	
	\section*{Acknowledgements}
	This work was partially supported by the Program for Guangdong Introducing Innovative and Entrepreneurial Teams (No.: 2017ZT07X183), the National Natural Science Foundation of China (No.: 61771201, 61902131), and the Guangdong R$\&$D key project of China (No.: 2019B010155001).
	

	{\small
		\bibliographystyle{ieee_fullname}
		\bibliography{egbib}

\begin{thebibliography}{10}\itemsep=-1pt

\bibitem{PCM_RegRecT}
I. Achituve, H. Maron, and G. Chechik.
\newblock Self-supervised learning for domain adaptation on point clouds.
\newblock In {\em Proceedings of the IEEE/CVF Winter Conference on Applications
  of Computer Vision}, pages 123--133, 2021.

\bibitem{CloserLookDN}
D. Arpit, S. Jastrz{{e}}bski, N. Ballas, D. Krueger, E. Bengio, M.~S. Kanwal,
  T. Maharaj, A. Fischer, A. Courville, Y. Bengio, and S. Lacoste-Julien.
\newblock A closer look at memorization in deep networks.
\newblock In {\em Int. Conf. Mach. Learn.}, pages 233--242, 2017.

\bibitem{da_theory1}
S. Ben-David, John B., K. Crammer, and F. Pereira.
\newblock Analysis of representations for domain adaptation.
\newblock In {\em Adv. Neural Inform. Process. Syst.}, pages 137--144, 2007.

\bibitem{da_theory2}
S. Ben-David, J. Blitzer, K. Crammer, A. Kulesza, F. Pereira, and J.~W.
  Vaughan.
\newblock A theory of learning from different domains.
\newblock {\em Mach. Learn.}, 79:151--175, 2010.

\bibitem{Shapenet}
A.~X Chang, T. Funkhouser, L. Guibas, P. Hanrahan, Q. Huang, Z. Li, S.
  Savarese, M. Savva, S. Song, H. Su, et~al.
\newblock Shapenet: An information-rich 3d model repository.
\newblock {\em ArXiv}, 1512.03012, 2015.

\bibitem{Clusternet}
C. Chen, G. Li, R. Xu, T. Chen, M. Wang, and L. Lin.
\newblock Clusternet: Deep hierarchical cluster network with rigorously
  rotation-invariant representation for point cloud analysis.
\newblock In {\em IEEE Conf. Comput. Vis. Pattern Recog.}, 2019.

\bibitem{bsp}
X. Chen, S. Wang, M. Long, and J. Wang.
\newblock Transferability vs. discriminability: Batch spectral penalization for
  adversarial domain adaptation.
\newblock In {\em Int. Conf. Mach. Learn.}, volume~97, pages 1081--1090, 2019.

\bibitem{SphericalC}
T. Cohen, M. Geiger, Jonas K{\"o}hler, and M. Welling.
\newblock Spherical cnns.
\newblock {\em ArXiv}, 2018.

\bibitem{Scannet}
A. Dai, A.~X Chang, M. Savva, M. Halber, T. Funkhouser, and M. Nie{\ss}ner.
\newblock Scannet: Richly-annotated 3d reconstructions of indoor scenes.
\newblock In {\em IEEE Conf. Comput. Vis. Pattern Recog.}, pages 5828--5839,
  2017.

\bibitem{cat}
Z. {Deng}, Y. {Luo}, and J. {Zhu}.
\newblock Cluster alignment with a teacher for unsupervised domain adaptation.
\newblock In {\em Int. Conf. Comput. Vis.}, pages 9943--9952, 2019.

\bibitem{StructuralRR}
Y. Duan, Y. Zheng, J. Lu, J. Zhou, and Q. Tian.
\newblock Structural relational reasoning of point clouds.
\newblock {\em IEEE Conf. Comput. Vis. Pattern Recog.}, 2019.

\bibitem{LearningSE}
C. Esteves, C. Allen-Blanchette, A. Makadia, and K. Daniilidis.
\newblock Learning so(3) equivariant representations with spherical cnns.
\newblock In {\em Eur. Conf. Comput. Vis.}, 2018.

\bibitem{dann}
Y. Ganin, E. Ustinova, H. Ajakan, P. Germain, H. Larochelle, F. Laviolette, M.
  Marchand, and V. Lempitsky.
\newblock Domain-adversarial training of neural networks.
\newblock {\em J. Mach. Learn. Res.}, 17:2096--2030, 2016.

\bibitem{gans}
I. Goodfellow, J. Pouget-Abadie, M. Mirza, B. Xu, D. Warde-Farley, S. Ozair, A.
  Courville, and Y. Bengio.
\newblock Generative adversarial nets.
\newblock In {\em Adv. Neural Inform. Process. Syst.}, pages 2672--2680, 2014.

\bibitem{can}
G. {Kang}, L. {Jiang}, Y. {Yang}, and A.~G. {Hauptmann}.
\newblock Contrastive adaptation network for unsupervised domain adaptation.
\newblock In {\em IEEE Conf. Comput. Vis. Pattern Recog.}, pages 4888--4897,
  2019.

\bibitem{adam}
D.~P Kingma and J. Ba.
\newblock Adam: A method for stochastic optimization.
\newblock {\em ArXiv}, 1412.6980, 2014.

\bibitem{swd}
C. {Lee}, T. {Batra}, M.~H. {Baig}, and D. {Ulbricht}.
\newblock Sliced wasserstein discrepancy for unsupervised domain adaptation.
\newblock In {\em IEEE Conf. Comput. Vis. Pattern Recog.}, pages 10277--10287,
  2019.

\bibitem{pseudo_label}
D.-H. Lee.
\newblock Pseudo-label : The simple and efficient semi-supervised learning
  method for deep neural networks.
\newblock In {\em ICML Workshop on Challenges in Representation Learning
  (WREPL)}, 2013.

\bibitem{SO_Net}
J. Li, B.~M. Chen, and G.~H. Lee.
\newblock So-net: Self-organizing network for point cloud analysis.
\newblock In {\em IEEE Conf. Comput. Vis. Pattern Recog.}, 2018.

\bibitem{PointCNN}
Y. Li, R. Bu, M. Sun, W. Wu, X. Di, and B. Chen.
\newblock Pointcnn: Convolution on x-transformed points.
\newblock In {\em Adv. Neural Inform. Process. Syst.}, pages 820--830, 2018.

\bibitem{cad}
J Lin, X Shi, Y Gao, K Chen, and K Jia.
\newblock Cad-pu: A curvature-adaptive deep learning solution for point set
  upsampling.
\newblock {\em arXiv preprint arXiv:2009.04660}, 2020.

\bibitem{SSL_Survey}
X. Liu, F. Zhang, Z. Hou, Z. Wang, L. Mian, J. Zhang, and J. Tang.
\newblock Self-supervised learning: Generative or contrastive.
\newblock {\em ArXiv}, 2006.08218, 2020.

\bibitem{dan}
M. {Long}, Y. {Cao}, Z. {Cao}, J. {Wang}, and M.~I. {Jordan}.
\newblock Transferable representation learning with deep adaptation networks.
\newblock {\em IEEE Trans. Pattern Anal. Mach. Intell.}, 41:3071--3085, 2019.

\bibitem{t-SNE}
L.~van~der Maaten and G. Hinton.
\newblock Visualizing data using t-sne.
\newblock {\em J. Mach. Learn. Res.}, 9:2579--2605, 2008.

\bibitem{transfer_learning_survey}
S.~J. {Pan} and Q. {Yang}.
\newblock A survey on transfer learning.
\newblock {\em IEEE Trans. Knowl. Data Eng.}, 22:1345--1359, 2010.

\bibitem{tpn}
Y. {Pan}, T. {Yao}, Y. {Li}, Y. {Wang}, C. {Ngo}, and T. {Mei}.
\newblock Transferrable prototypical networks for unsupervised domain
  adaptation.
\newblock In {\em IEEE Conf. Comput. Vis. Pattern Recog.}, pages 2234--2242,
  2019.

\bibitem{SimNet}
P.~O. {Pinheiro}.
\newblock Unsupervised domain adaptation with similarity learning.
\newblock In {\em IEEE Conf. Comput. Vis. Pattern Recog.}, pages 8004--8013,
  2018.

\bibitem{SSL_RotPred_PC}
O. Poursaeed, T. Jiang, Q. Qiao, N. Xu, and V. Kim.
\newblock Self-supervised learning of point clouds via orientation estimation.
\newblock In {\em 3DV}, 2020.

\bibitem{Pointnet}
C.~R Qi, H. Su, K. Mo, and L.~J Guibas.
\newblock Pointnet: Deep learning on point sets for 3d classification and
  segmentation.
\newblock In {\em IEEE Conf. Comput. Vis. Pattern Recog.}, pages 652--660,
  2017.

\bibitem{Pointnet++}
C.~R. Qi, L. Yi, H. Su, and L.~J Guibas.
\newblock Pointnet++: Deep hierarchical feature learning on point sets in a
  metric space.
\newblock In {\em Adv. Neural Inform. Process. Syst.}, pages 5099--5108, 2017.

\bibitem{PointDAN}
C. Qin, H. You, L. Wang, C.-C.~J. Kuo, and Y. Fu.
\newblock Pointdan: A multi-scale 3d domain adaption network for point cloud
  representation.
\newblock In {\em Adv. Neural Inform. Process. Syst.}, pages 7192--7203, 2019.

\bibitem{BeyondSW}
A. Rozantsev, M. Salzmann, and P. Fua.
\newblock Beyond sharing weights for deep domain adaptation.
\newblock {\em IEEE Trans. Pattern Anal. Mach. Intell.}, 41:801--814, 2019.

\bibitem{adr}
K. Saito, Y. Ushiku, T. Harada, and K. Saenko.
\newblock Adversarial dropout regularization.
\newblock In {\em Int. Conf. Learn. Represent.}, 2018.

\bibitem{mcd}
K. {Saito}, K. {Watanabe}, Y. {Ushiku}, and T. {Harada}.
\newblock Maximum classifier discrepancy for unsupervised domain adaptation.
\newblock In {\em IEEE Conf. Comput. Vis. Pattern Recog.}, pages 3723--3732,
  2018.

\bibitem{SSL_PointSeg_PC}
Jonathan Sauder and Bjarne Sievers.
\newblock Self-supervised deep learning on point clouds by reconstructing
  space.
\newblock In {\em Adv. Neural Inform. Process. Syst.}, pages 12962--12972,
  2019.

\bibitem{dirt_t}
R. Shu, H. Bui, H. Narui, and S. Ermon.
\newblock A {DIRT}-t approach to unsupervised domain adaptation.
\newblock In {\em Int. Conf. Learn. Represent.}, 2018.

\bibitem{SSL_UDA_Images}
Y. Sun, E. Tzeng, T. Darrell, and Alexei~A. Efros.
\newblock Unsupervised domain adaptation through self-supervision.
\newblock {\em ArXiv}, 1909.11825, 2019.

\bibitem{tang2020improving}
L. Tang, K. Chen, C. Wu, Y. Hong, K. Jia, and Z. Yang.
\newblock Improving semantic analysis on point clouds via auxiliary supervision
  of local geometric priors.
\newblock {\em ArXiv}, 2001.04803, 2020.

\bibitem{KPconv}
H. Thomas, C.~R Qi, J. Deschaud, B. Marcotegui, F. Goulette, and L.~J Guibas.
\newblock Kpconv: Flexible and deformable convolution for point clouds.
\newblock In {\em Int. Conf. Comput. Vis.}, pages 6411--6420, 2019.

\bibitem{adda}
E. {Tzeng}, J. {Hoffman}, K. {Saenko}, and T. {Darrell}.
\newblock Adversarial discriminative domain adaptation.
\newblock In {\em IEEE Conf. Comput. Vis. Pattern Recog.}, pages 2962--2971,
  2017.

\bibitem{Modelnet}
K.~V. Vishwanath, Diwaker Gupta, Amin Vahdat, and Ken Yocum.
\newblock Modelnet: Towards a datacenter emulation environment.
\newblock In {\em 2009 IEEE Ninth International Conference on Peer-to-Peer
  Computing}, pages 81--82, 2009.

\bibitem{DGCNN}
Y. Wang, Y. Sun, Z. Liu, S.~E Sarma, M.~M Bronstein, and J.~M Solomon.
\newblock Dynamic graph cnn for learning on point clouds.
\newblock {\em Acm Transactions On Graphics (tog)}, 38(5):1--12, 2019.

\bibitem{noisy_student}
Q. Xie, M.-T. Luong, E. Hovy, and Q.~V Le.
\newblock Self-training with noisy student improves imagenet classification.
\newblock In {\em IEEE Conf. Comput. Vis. Pattern Recog.}, pages 10687--10698,
  2020.

\bibitem{mstn}
S. Xie, Z. Zheng, L. Chen, and C. Chen.
\newblock Learning semantic representations for unsupervised domain adaptation.
\newblock In {\em Int. Conf. Mach. Learn.}, pages 5423--5432, 2018.

\bibitem{PointASNLRP}
X. Yan, C. Zheng, Z. Li, S. Wang, and S. Cui.
\newblock Pointasnl: Robust point clouds processing using nonlocal neural
  networks with adaptive sampling.
\newblock {\em IEEE Conf. Comput. Vis. Pattern Recog.}, 2020.

\bibitem{ModelingPC}
J. Yang, Q. Zhang, B. Ni, L. Li, J. Liu, M. Zhou, and Q. Tian.
\newblock Modeling point clouds with self-attention and gumbel subset sampling.
\newblock {\em IEEE Conf. Comput. Vis. Pattern Recog.}, 2019.

\bibitem{mixup}
H. Zhang, M. Cisse, Y.N. Dauphin, and D. Lopez-Paz.
\newblock mixup: Beyond empirical risk minimization.
\newblock {\em arXiv preprint arXiv:1710.09412}, 2017.

\bibitem{Pointweb}
H. Zhao, Li Jiang, C. Fu, and J. Jia.
\newblock {PointWeb}: Enhancing local neighborhood features for point cloud
  processing.
\newblock In {\em IEEE Conf. Comput. Vis. Pattern Recog.}, 2019.

\bibitem{cbst}
Y. Zou, Z. Yu, B.V.K.~Vijaya Kumar, and J. Wang.
\newblock Unsupervised domain adaptation for semantic segmentation via
  class-balanced self-training.
\newblock In {\em Eur. Conf. Comput. Vis.}, pages 289--305, 2018.

\end{thebibliography}
	}
	
\end{document}